%% file: main.tex
\documentclass[10pt,journal,compsoc]{IEEEtran}

\ifCLASSOPTIONcompsoc
  \usepackage[nocompress]{cite}
\else
  \usepackage{cite}
\fi

\ifCLASSINFOpdf
\else
\fi

\usepackage{epsfig}
\usepackage{graphicx}
\usepackage{amsmath}
\usepackage{amssymb}
\usepackage{enumitem}
\usepackage{ragged2e}
\usepackage[table]{xcolor}
\usepackage{textcomp}
\usepackage{amssymb}%
\usepackage{pifont}%
\usepackage{stfloats}

\usepackage[dvipsnames]{xcolor}
\usepackage[
    pagebackref=true,
    breaklinks=true,
    colorlinks=true,
    bookmarks=false,
    urlcolor=NavyBlue,
    citecolor=NavyBlue
]{hyperref}
\usepackage{gensymb,graphics,inputenc}
\usepackage{caption}
\usepackage{subcaption}
\usepackage[linesnumbered,algo2e,boxed]{algorithm2e}
\graphicspath{{Figure/}}
\usepackage[figuresright]{rotating}
\usepackage{amsmath,amssymb,textcomp,multirow,upgreek}
\usepackage{booktabs}
\usepackage{color,mdwlist}

\usepackage[hang,flushmargin]{footmisc}

\usepackage{graphicx,fontawesome}
\graphicspath{{Figures/}}

\input{commands}

\usepackage{arydshln}

\begin{document}
\title{Track-On2:\\Enhancing Online Point Tracking with Memory}

\author{\IEEEauthorblockN{Görkay Aydemir \quad
        Weidi Xie~\textsuperscript{$\dagger$} \quad
        Fatma Güney~\textsuperscript{$\dagger$}
	}
	\thanks{\par Görkay Aydemir is the corresponding author (gaydemir23@ku.edu.tr). \protect
            \par Görkay Aydemir and Fatma Güney are with the Koç University, Istanbul, Turkey. Weidi Xie is with the School of Artificial Intelligence, Shanghai Jiao Tong University, China. \par 
            \textsuperscript{$\dagger$} denotes equal supervision \protect}
}

\markboth{IEEE Transactions on Pattern Analysis and Machine Intelligence}%
{Shell \MakeLowercase{\textit{et al.}}: Bare Demo of IEEEtran.cls for Computer Society Journals}

\IEEEtitleabstractindextext{%
\input{sec/00-abstract}

}

\maketitle

\IEEEdisplaynontitleabstractindextext

\input{sec/01-intro}
\input{sec/02-rw}
\input{sec/03-methodology}
\input{sec/04-exp}

\input{sec/05-conclusion}
\input{sec/07-ack}

\small
\bibliographystyle{IEEEtran}
\bibliography{bibliography_long, ref}

\input{sec/06-bio}

\end{document}

%% file: commands.tex
\newcommand{\bI}{\mathbf{I}}

\newcommand{\bs}{\mathbf{s}}
\newcommand{\bM}{\mathbf{M}}

\newcommand{\bv}{\mathbf{v}}
\newcommand{\bq}{\mathbf{q}}

\newcommand{\bp}{\mathbf{p}}

\newcommand{\bu}{\mathbf{u}}

\newcommand{\bC}{\mathbf{C}}

\newcommand{\bff}{\mathbf{f}}

\newcommand{\bo}{\mathbf{o}}

\newcommand{\bc}{\mathbf{c}}

\newcommand{\bh}{\mathbf{h}}

\newcommand{\nR}{\mathbb{R}}

\newcommand{\cL}{\mathcal{L}}

\newcommand{\cQ}{\mathcal{Q}}
\newcommand{\cV}{\mathcal{V}}

\newcommand{\figref}[1]{Fig.~\ref{#1}}
\newcommand{\secref}[1]{Sec.~\ref{#1}}

\newcommand{\tabref}[1]{Table~\ref{#1}}
\newcommand{\repro}[1]{#1\textsuperscript{$\dagger$}}

\makeatletter
\DeclareRobustCommand\onedot{\futurelet\@let@token\@onedot}
\def\@onedot{\ifx\@let@token.\else.\null\fi\xspace}
\def\eg{{\em e.g}\onedot}

\def\ie{{\em i.e}\onedot} 
 
 \def\vs{vs\onedot}

\makeatother

\newcommand{\boldparagraph}[1]{\noindent{\bf #1:} }

\newcommand{\ga}[1]{{#1}}

\newcommand{\cmark}{\ding{51}}%
\newcommand{\xmark}{\ding{55}}%

\newcommand{\deltaavg}{$\delta^{x}_{avg}$}

\newcommand{\up}{$\uparrow$}

\definecolor{softgreen}{RGB}{247,253,245}   %
\definecolor{softblue}{RGB}{244,247,252}    %
\definecolor{offwhite}{RGB}{253,253,253}    %

%% file: sec/00-abstract.tex
\justify

\begin{abstract}

In this paper, we consider the problem of long-term point tracking, which requires consistent identification of points across video frames under significant appearance changes, motion, and occlusion. We target the online setting, {\em i.e.}, tracking points frame-by-frame, making it suitable for real-time and streaming applications. %
We extend our prior model Track-On into \textbf{Track-On2}, a simple and efficient transformer-based model for online long-term tracking. Track-On2 improves both performance and efficiency through architectural refinements, more effective use of memory, and improved synthetic training strategies. Unlike prior approaches that rely on full-sequence access or iterative updates, our model processes frames causally and maintains temporal coherence via a memory mechanism, which is key to handling drift and occlusions without requiring future frames. At inference, we perform coarse patch-level classification followed by refinement.
Beyond architecture, we systematically study synthetic training setups and their impact on memory behavior, showing how they shape temporal robustness over long sequences. Through comprehensive experiments, Track-On2 achieves state-of-the-art results across five synthetic and real-world benchmarks, surpassing prior online trackers and even strong offline methods that exploit bidirectional context. These results highlight the effectiveness of causal, memory-based architectures trained purely on synthetic data as scalable solutions for real-world point tracking. Project page: \url{https://kuis-ai.github.io/track_on2}.

\end{abstract}

%% file: sec/01-intro.tex
\section{Introduction}

Motion estimation is one of the core challenges in computer vision, 
with applications spanning video compression \cite{Jasinschi1998JFI}, video stabilization \cite{Battiato2007ICIAP, Lee2009ICCV}, and augmented reality \cite{Marchand2015VCG}. The objective is to track physical points across video frames accurately. A widely used solution for motion estimation is optical flow, which estimates pixel-level correspondences between adjacent frames. 
In principle, long-term motion estimation can be achieved by chaining together these frame-by-frame estimations.

Recent advances in optical flow, such as PWC-Net \cite{Sun2018CVPR} and RAFT \cite{Teed2020ECCV}, have improved accuracy for short-term motion estimation. 
However, the inherent limitations of chaining flow estimations remain a challenge, namely error accumulation and the difficulty of handling occlusions. To address long-term motion estimation, \cite{Sand2008IJCV} explicitly introduced the concept of pixel tracking, a paradigm shift that focuses on tracking individual points across a video, rather than relying solely on pairwise frame correspondences. This concept, often referred to as ``particle video” has been revisited in recent deep learning methods, for example, PIPs \cite{Harley2022ECCV} and TAPIR \cite{Doersch2023ICCV}, which leverage dense cost volumes, iterative optimization, and learned appearance updates to track points through time.

Generally speaking, the existing methods for long-term point tracking face two major limitations. First, they primarily rely on offline processing, where the entire video or a large window of frames is processed at once. This allows models to use both past and future frames to improve predictions but inherently limits their applicability in real-time scenarios \cite{Karaev2024ECCV, Harley2022ECCV}. Second, these approaches struggle with scalability, as they often require full attention computation across all frames, leading to significant memory overhead, especially for long videos or large frame windows. These limitations hinder their use in real-world applications, like robotics or augmented reality, where efficient and online processing of streaming video is crucial.

\begin{figure}[t]
    \centering
    \includegraphics[width=1\linewidth]{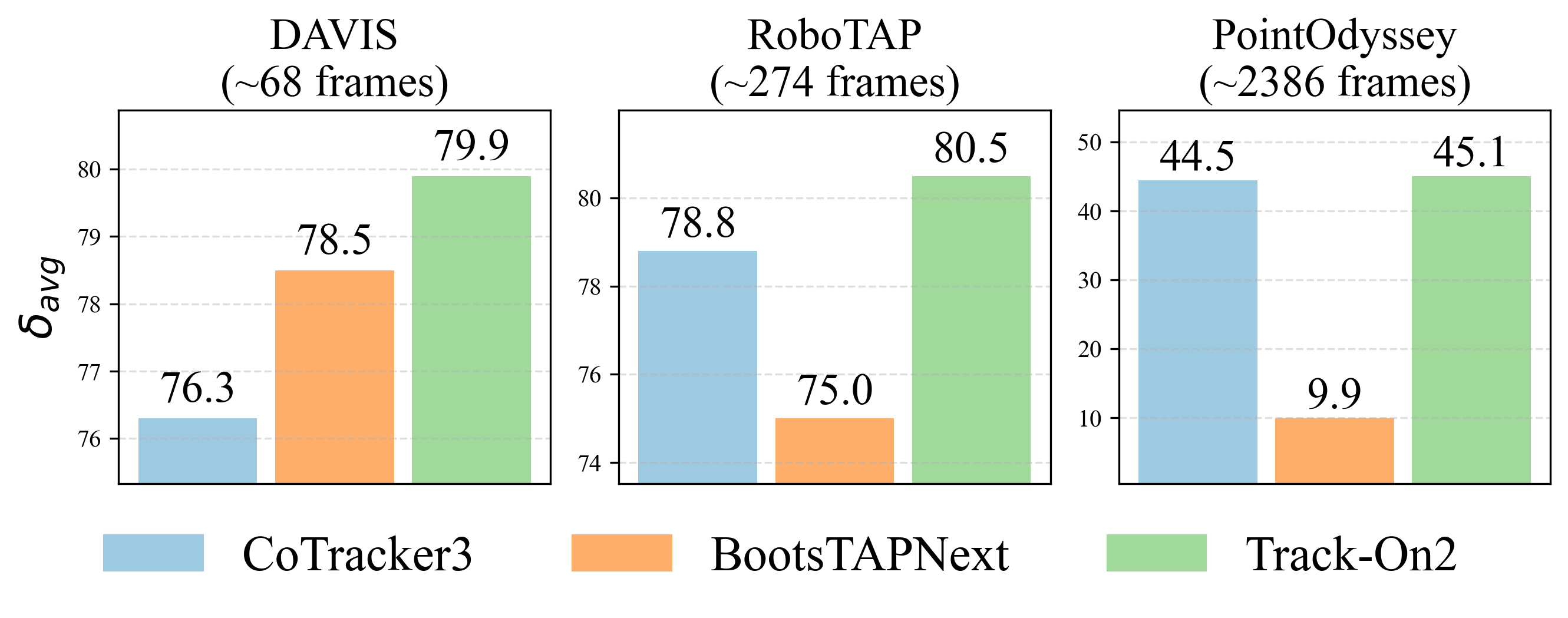}
    \caption{
    \textbf{Comparison with prior state-of-the-art.} We report \deltaavg on three representative benchmarks: short real videos (DAVIS), mid-length robotic sequences (RoboTAP), and very long synthetic videos (PointOdyssey). CoTracker3~\cite{Karaev2024ARXIV} and BootsTAPNext~\cite{Zholus2025ICCV} are fine-tuned on real-world data, which boosts their performance. Despite relying only on synthetic training, \textbf{Track-On2} achieves higher accuracy.
    }
    \vspace{-10pt}
    \label{fig:sota_hist}
\end{figure}

\begin{figure*}[t]
    \centering
    \includegraphics[width=.9\linewidth]{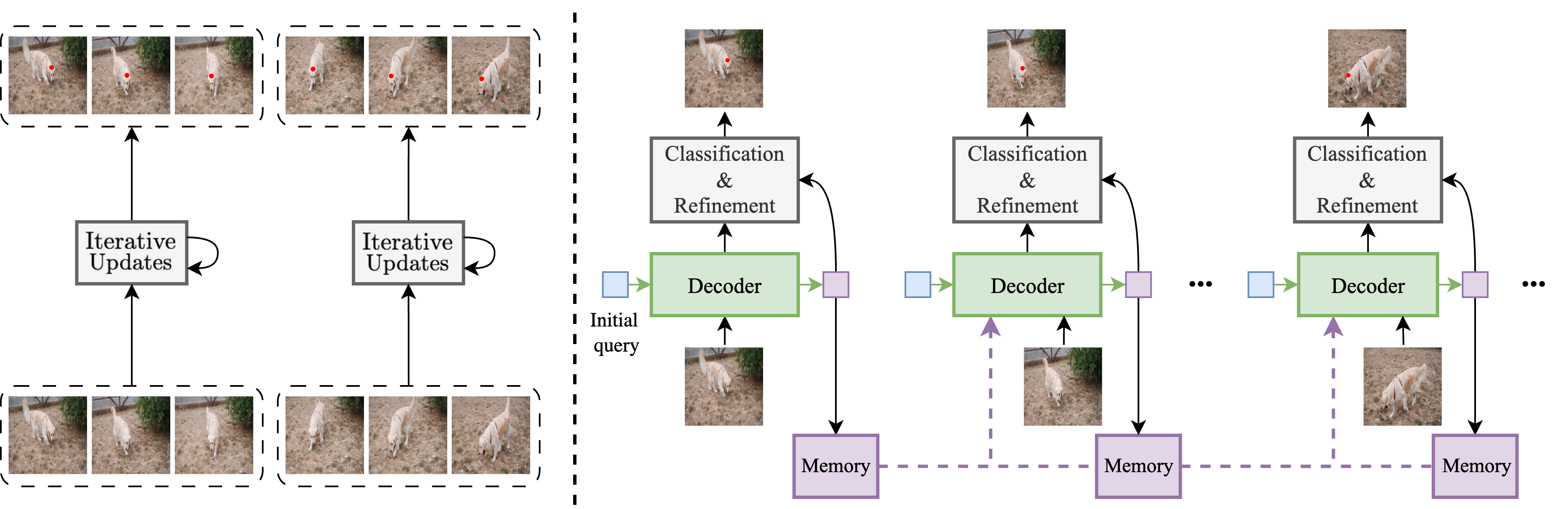}
    \caption{
    \textbf{Offline \vs Online Point Tracking.} We propose an online model, tracking points frame-by-frame (\textbf{right}), unlike the dominant offline paradigm where models require access to all frames within a sliding window or the entire video (\textbf{left}). To capture temporal information, we introduce a memory, storing contextual information from previous states of the point. The blue box denotes the initial query sampled from the first-frame point location, and the purple boxes represent the decoded queries at each timestep, conditioned on both the initial sample and the evolving memory.}
    \label{fig:teaser}
\end{figure*}

In our ICLR’25 paper, we introduced \textbf{Track-On}, a simple transformer-based model for online long-term point tracking~(\figref{fig:teaser}, right). The task is to process video frames sequentially, without access to the future, and still maintain consistent point identities across long sequences. Specifically, the points of interest are treated as queries in the decoder, which attend to the current frame features to update their representation. At training time, rather than directly regressing coordinates as in prior work, Track-On adopts a classification-first formulation: each query scores candidate patches via embedding similarity and is supervised with a similarity-based classification loss. 
At inference time, the model performs coarse patch-level classification, applies a lightweight re-ranking step to suppress distractors, and finally regresses a sub-patch offset for precise localization. To enable consistent tracking without future frames, Track-On maintains two memory modules that propagate contextual cues across time. These memory structures carry history forward, allowing the model to resolve ambiguities caused by appearance changes or occlusions. Importantly, the memory design supports inference-time memory extension (IME), which allows the model to scale beyond training-time sequence lengths without retraining.

This work presents \textbf{Track-On2}, with architectural simplifications and deeper empirical analysis over the ICLR25 version. Track-On2 streamlines the dual-memory design into a single expandable memory, building on the inference-time memory extension (IME) introduced in Track-On, and replaces pooled single-scale maps with multi-scale features from ViT-Adapter\cite{Chen2023ICLR}. These changes improve both speed and accuracy, enabling real-time operation ($>$30 FPS) with a low memory footprint. 

\begin{figure}[!b]
    \centering
    \includegraphics[width=1\linewidth]{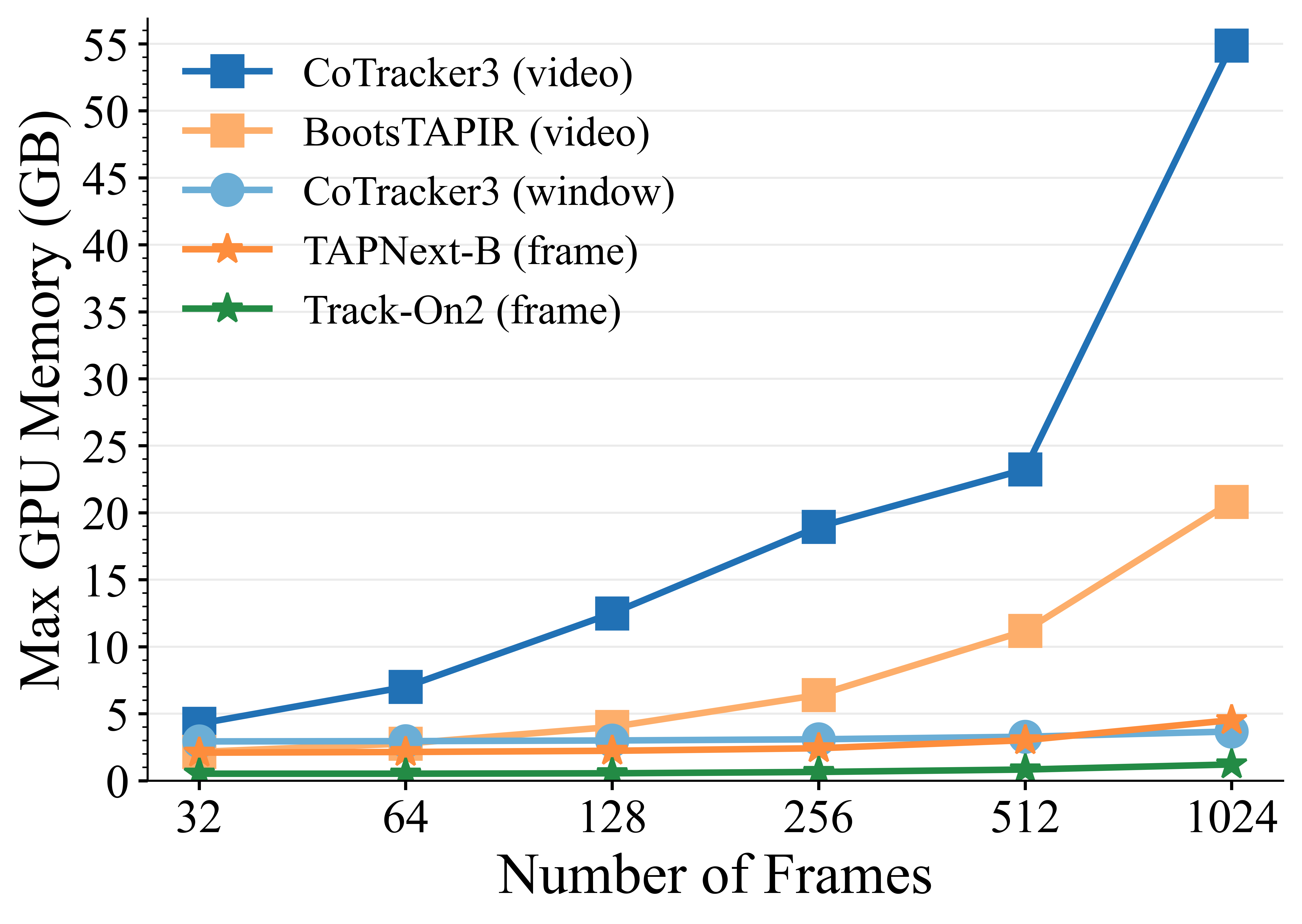}
    \caption{
    \textbf{Max GPU memory usage with increasing video length.} We plot the maximum GPU memory usage of state-of-the-art models when tracking 256 points in videos of varying lengths. Models are grouped into three categories based on their input processing strategy: 
    Video, window, and frame. It is observed that video-level models scale poorly and quickly run out of memory on longer sequences, while others remain efficient across all lengths
    }
    \label{fig:gpu_mem_vs_t}
\end{figure}

Beyond architecture, we train the model on longer clips, thereby increasing its effective memory capacity and improving temporal robustness. 
We further provide systematic empirical studies into how memory interacts with training supervision: in particular, we show that the length of training videos is the dominant factor governing memory behavior and generalization to long horizons, while frame sampling strategy and memory capacity (both during training and at inference, \ie IME) play complementary roles. Unlike Track-On, where memory acted as a fixed module, Track-On2 demonstrates that temporal reasoning emerges from the relation between memory design and training clip length. These insights clarify memory’s role as a learned mechanism and demonstrate that, when designed carefully, synthetic pretraining alone can yield state-of-the-art results. As shown in~\figref{fig:sota_hist}, Track-On2 sustains top performance across benchmarks, generalizing from short to long sequences and from synthetic domains to real-world videos, where other models often struggle. Its frame-by-frame inference ensures scalability: increasing the number of frames does not affect memory usage (\figref{fig:gpu_mem_vs_t}), unlike full-sequence models whose cost grows with sequence length.

In summary, our contributions are as follows:
(i) We propose an efficient architecture for point tracking, termed as Track-On2, with a single expandable memory and stronger multi-scale features, achieving high FPS with reduced inference latency and memory footprint, while improving accuracy;
(ii) We conduct extensive ablations and show that training video length is the dominant factor that shapes memory behavior and long-term scalability, with complementary effects from sampling strategies, memory configuration, and inference time memory extension;
(iii) We evaluate Track-On2 on five widely used benchmarks, covering both synthetic and real-world datasets, and show that it consistently matches or surpasses prior online and offline models, including those fine-tuned on real videos, despite being trained solely on synthetic training and strictly causal inference.

%% file: sec/02-rw.tex
\section{Related Work}

\subsection{Point Tracking}
Point tracking presents significant challenges, particularly for long-term tracking, where maintaining consistent tracking through occlusions is difficult. PIPs~\cite{Harley2022ECCV} was one of the first approaches to address this by predicting motion through iterative updates within temporal windows. TAP-Vid~\cite{Doersch2022NeurIPS} initiated a benchmark for evaluation. TAPIR~\cite{Doersch2023ICCV} improved upon PIPs by refining initialization and incorporating depthwise convolutions to enhance temporal accuracy. In contrast, CoTracker~\cite{Karaev2024ECCV} introduced a novel approach by jointly tracking multiple points, exploiting spatial correlations between points via factorized transformers. Differently, TAPTR~\cite{Li2024ECCV} adopted a design inspired by DETR~\cite{Carion2020ECCV, Zhu2021ICLR}, drawing parallels between object detection and point tracking. TAPTRv2~\cite{Li2024NeurIPS}, the successor to TAPTR, builds on its predecessor by incorporating offsets predicted by the deformable attention module. While these models calculate point-to-region similarity for correlation, LocoTrack~\cite{Cho2024ECCV} introduced a region-to-region similarity approach to address ambiguities in matching. CoTracker3~\cite{Karaev2024ARXIV} combined the region-to-region similarity method from LocoTrack with the original CoTracker architecture. %
\textbf{Track-On}~\cite{Aydemir2025ICLR} introduced a distinct classification-first formulation: queries perform patch-level classification followed by refinement. In this work, we follow the same classification-first formulation.

\vspace{3pt}\boldparagraph{Online Point Tracking}
{While the dominant approach in point tracking assumes offline access to all frames, either via full sequences~\cite{Doersch2023ICCV, Doersch2024ARXIV} or within sliding windows~\cite{Karaev2024ECCV}, a parallel line of research focuses on causal models that process video sequentially. %
MFT~\cite{Neoral2024WACV}, which extends optical flow to long-term scenarios, can be adapted for online point tracking tasks, although it does not belong to the point tracking family. 
Recently, TAPNext~\cite{Zholus2025ICCV} proposed using state-space models~\cite{Team2024arXiv} to capture temporal dependencies, offering a new way to perform point tracking in a fully online fashion.
Among other point tracking approaches, models with online variants~\cite{Doersch2024ARXIV, Doersch2023ICCV} are re-trained with a temporally causal mask to process frames sequentially on a frame-by-frame basis, despite being originally designed for offline tracking.
In contrast, Track-On was designed explicitly for online tracking, introducing memory modules to stabilize long sequences. Track-On2 continues this line but reduces the overhead of the dual-memory design by using a single expandable memory, enabling more efficient online tracking without sacrificing robustness.

\vspace{3pt}\boldparagraph{Fine-tuning on Real-world Videos}
Due to the scarcity of manually annotated point tracks, most existing models are pretrained on the synthetic TAP-Vid-Kubric dataset~\cite{Doersch2022NeurIPS}. Several recent efforts aim to improve real-world performance via domain adaptation: BootsTAPIR~\cite{Doersch2024ARXIV} uses student-teacher distillation on large-scale unlabeled video data, while CoTracker3~\cite{Karaev2024ARXIV} further leverages pseudo-labeled real-world videos during training. Despite these advances, all such methods still rely heavily on synthetic pretraining, not only as initialization but also as a source of teacher models and supervision signals.
Track-On2 focuses exclusively on synthetic supervision, systematically analyzing how to maximize its effectiveness. %

\vspace{-10pt} \subsection{Causal Processing in Videos}
Online or temporally causal models process frames sequentially, relying only on current and past inputs without future access. This contrasts with the dominant clip-based paradigm in point tracking. Causal processing is particularly useful for streaming~\cite{Yang2022CVPRb, Zhou2024CVPR}, embodied perception~\cite{Yao2019IROS}, and long videos~\cite{Zhang2024ARXIV, Xu2021NeurIPS}, where activation caching and scalability are important. As a result, causal modeling has been explored across tasks like pose estimation~\cite{Fan2021ICCV, Nie2019ICCV}, action detection~\cite{Xu2019ICCV, De2016ECCV, Kondratyuk2021CVPR, Eun2020CVPR, Yang2022CVPRa, Wang2021ICCV, Zhao2022ECCV, Xu2021NeurIPS, Chen2022CVPR}, object tracking~\cite{He2018CVPR, Wang2020ECCV}, video captioning~\cite{Zhou2024CVPR}, and video object segmentation~\cite{Cheng2022ECCV, Liang2020NeurIPS}.
Context propagation in causal models is typically handled via recurrent structures~\cite{De2016ECCV} or transformers with causal masks~\cite{Wang2021ICCV}, but both face limitations in memory and temporal range. To improve efficiency and capacity, memory modules have been proposed. For example, LSTR~\cite{Xu2021NeurIPS} separates long- and short-term memory for action detection, and XMem~\cite{Cheng2022ECCV} uses fine-grained memory in segmentation. \ga{In visual tracking, ODTrack~\cite{Zheng2024AAAI} reformulates tracking as online token propagation, enabling dense inter-frame association without image-pair matching. UM-ODTrack~\cite{Zheng2025TPAMI} extends this idea to a unified framework while maintaining temporally causal processing. Similar memory-augmented or token-based propagation strategies have also been explored in long-text modeling~\cite{Balazevic2024ICML} and video question answering~\cite{Zhang2021ICML}.} %

%% file: sec/03-methodology.tex
\vspace{-10pt} \section{Methodology}

\begin{figure*}[t]
    \centering
    \includegraphics[width=1\linewidth]{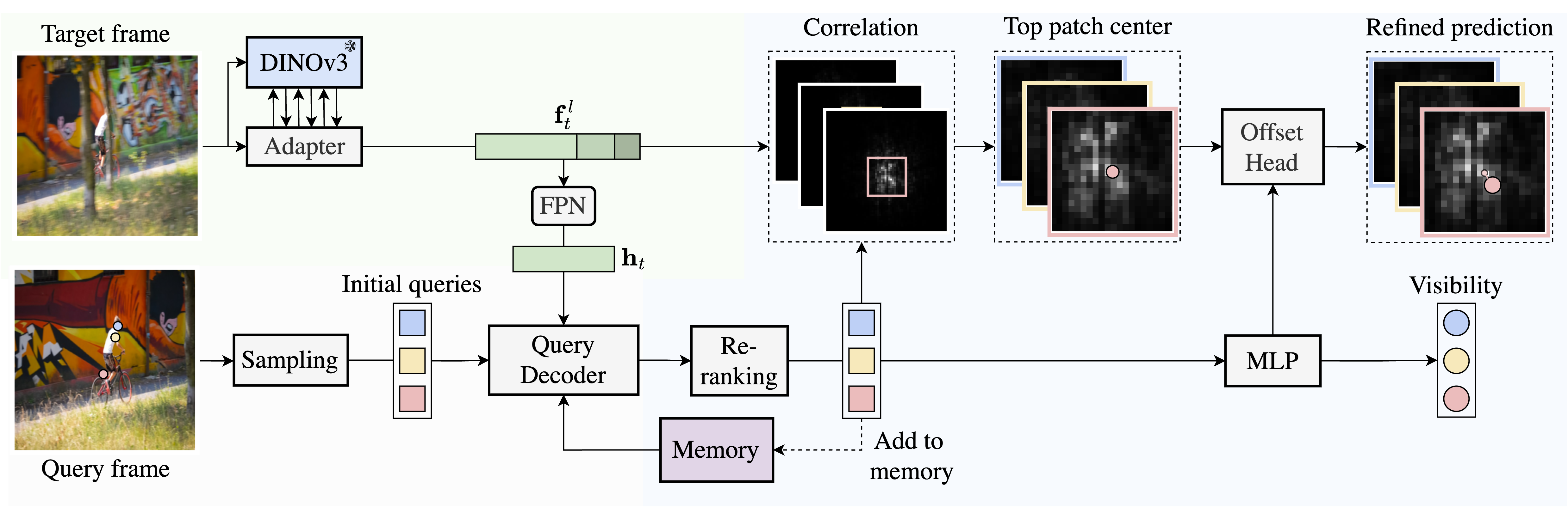}
    \caption{
    \textbf{Overview.} We introduce Track-On2, a transformer-based method for online, frame-by-frame point tracking. The pipeline consists of three stages:
    (i) \textbf{Visual Encoder}~(\raisebox{0pt}[0.9ex][0.4ex]{\colorbox{softgreen}{top-left}}), which extracts multi-scale features from each frame using a DINOv3-based ViT-Adapter and fuses them via an FPN;
    (ii) \textbf{Query Decoding} (bottom-left), where point queries attend to the current frame features and a persistent memory propagated from the previous frame;
    (iii) \textbf{Point Prediction}~(\raisebox{0pt}[0.9ex][0.4ex]{\colorbox{softblue}{right}}), which estimates correspondences in a coarse-to-fine manner.
    Decoded queries are refined by a re-ranking module that incorporates local information from candidate matches, and point locations are predicted by patch-level classification with sub-patch regression, alongside a lightweight visibility head. Refined queries are appended to memory for use in the next frame.
    $\Box$ denotes a point query, and \small{\textcircled{}} denotes a prediction.
    }
    \label{fig:main_fig}
\end{figure*}

\subsection{Problem Scenario}
Given an RGB video of $T$ frames, $\cV = \bigl\{\bI_1,~ \bI_2,~ \dots,~ \bI_T \bigr\} \in \nR^{T \times H \times W \times 3}$, 
and a set of $N$ predefined queries, 
$\cQ=  \bigl\{ (t^{1}, \bp^{1}),~(t^{2}, \bp^{2}),~ \dots,~ ~(t^{N}, \bp^{N}) \bigr\}\in \nR^{N \times 3}$, 
where each query point is specified by the start time and pixel's spatial location, our goal is to predict the correspondences $\hat{\bp}_t \in \nR^{N \times 2}$ and visibility $\hat{\bv}_t \in \{0, 1\}^{N}$ for all query points in an online manner, \ie using only frames up to the current target frame $t$. To address this problem, we propose a transformer-based point tracking model, $\Phi(\cdot;\Theta)$, that tracks points \textbf{frame-by-frame}, 
with a dynamic memory $\bM$ to propagate temporal information along the video sequence:
\begin{equation}
\hat{\bp}_t,~ \hat{\bv}_t,~ \bM_t = \Phi \left(\bI_t,~ \cQ,~ \bM_{t-1};~ \Theta  \right)
\end{equation}

In the following section, we aim to detail our model and introduce the memory module.

\subsection{Architecture}

Our model is based on transformer, consisting of three components, as illustrated in \figref{fig:main_fig}: \textbf{Visual Encoder}~(\ref{sec:vis_encoder}) is tasked to extract visual features of the video frame, and initialize the query points; \textbf{Query Decoder}~(\ref{sec:query_decoder}) enables the queried points to attend the target frame and the memory, to update their features; and \textbf{Point Prediction}~(\ref{sec:point_prediction}), to predict the positions of corresponding queried points in a coarse-to-fine manner.

\subsubsection{Visual Encoder}
\label{sec:vis_encoder}
We adopt a Vision Transformer (ViT) as visual backbone, specifically, DINOv3 ViT-S+~\cite{Simeoni2025arXiv}, and use ViT-Adapter~\cite{Chen2023ICLR} to obtain dense features at 4 different scales: 
\begin{equation}
\bff_t^l = \Phi_{\text{vis-enc}} \left(\bI_t\right) \in \nR^{\frac{H}{2^{l} \cdot S} \times \frac{W}{2^{l} \cdot S} \times D},  \quad  l \in \{0, 1, 2, 3 \}
\end{equation}
where $D$ denotes the feature dimension, and $S = 4$ refers to the stride. To fuse multi-scale features at the highest resolution, we fuse them using a simple Feature Pyramid Network (FPN)~\cite{Lin2017CVPR} $\Phi_\text{fpn}$:
\begin{equation}
\bh_t  = \Phi_\text{fpn} \left(\bff_t^{\{0, 1, 2, 3\}} \right) \in \nR^{\frac{H}{S} \times \frac{W}{S} \times D}
\end{equation}
Each scale is first encoded with a lightweight layer, and features are then progressively upsampled from coarse to fine and summed with higher-resolution features. This produces a unified representation that combines semantic context with spatial detail at the highest resolution.

\vspace{3pt}\boldparagraph{Query Initialization} 
To initialize the query features~($\bq^{init}$), 
we apply bilinear sampling to the feature map at the query location~$\left(\bp^{i}\right)$:
\begin{equation*}
\bq^{init} = \bigl\{\text{sample}(\bh_{t^{i}},~ \bp^{i})\bigr\}_{i = 1}^{N} \in \nR^{N \times D}
\end{equation*}
In practice, we initialize the query based on the features of the start frame $t^{i}$ for $i$-th query and propagate them to the subsequent frames.

\vspace{3pt}\boldparagraph{Differences from Track-On}
Compared to the original Track-On~\cite{Aydemir2025ICLR}, our visual encoder introduces two key changes. First, Track-On relied on a single-resolution feature map, limiting spatial flexibility. In Track-On2, we leverage the native multi-scale feature extraction capabilities of ViT-Adapter~\cite{Chen2023ICLR}, and fuse them using an FPN~\cite{Lin2017CVPR}, yielding a more expressive and high-resolution representation. Second, while Track-On used DINOv2 ViT-S as the backbone, we adopt DINOv3 ViT-S+\cite{Simeoni2025arXiv} for improved visual quality. For fair comparison, we also report results using DINOv2 ViT-S in our comparisons (\secref{sec:exp}).

\subsubsection{Query Decoder}
\label{sec:query_decoder}

\begin{figure}[t]
    \centering
    \includegraphics[width=1\linewidth]{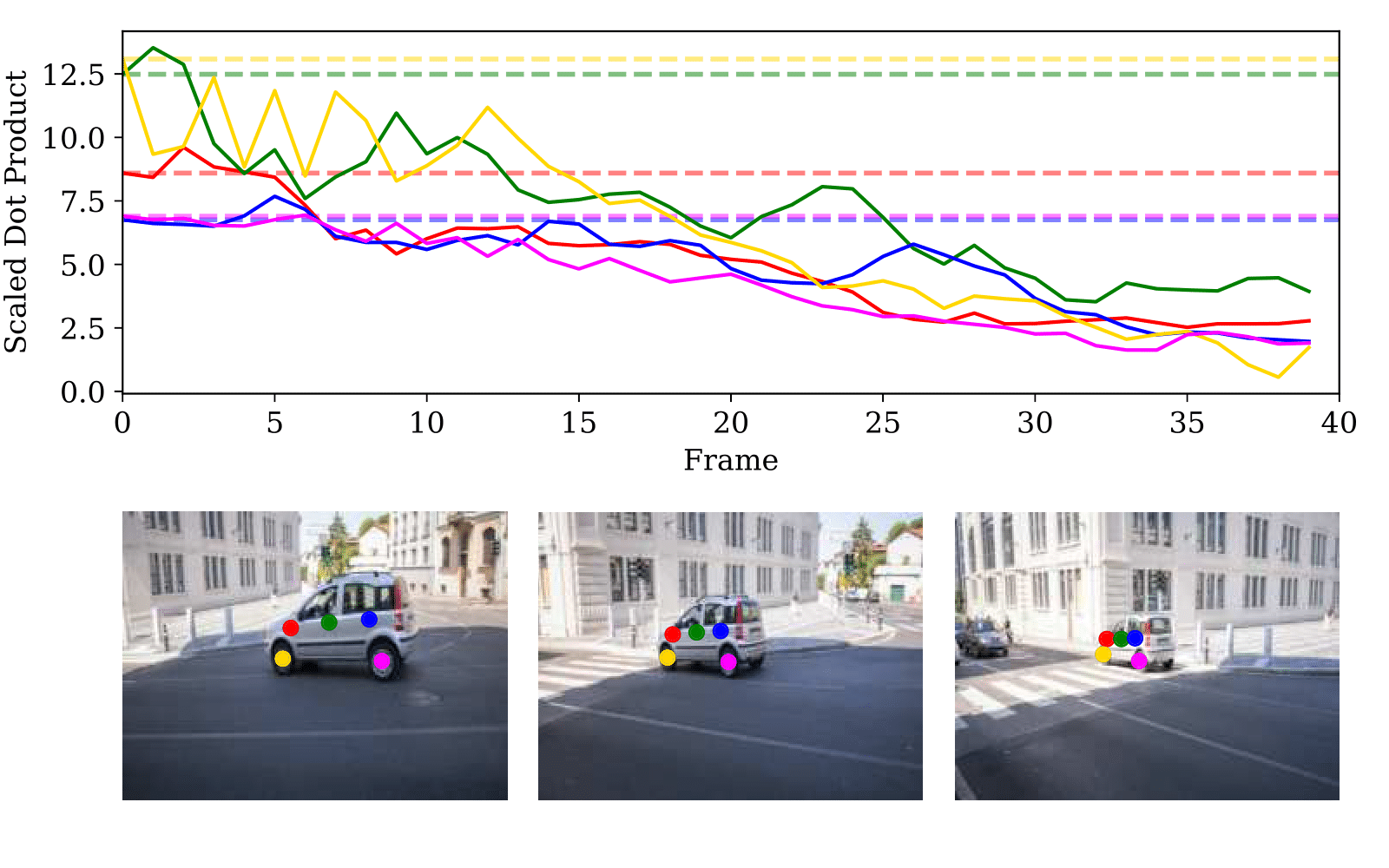}
    \caption{\textbf{Feature Drift.} %
    For the tracks shown below (start, middle, and final frames), the plot above illustrates the decreasing similarity between the features of the initial query and its correspondences over time, with the initial similarity indicated by horizontal dashed lines.
    }
    \label{fig:feature_drift}
\end{figure}

As point appearances evolve over time, similarity between the initial query and its true matches decays (\figref{fig:feature_drift}), a phenomenon known as feature drift. Relying solely on the initial query embedding thus yields increasingly inaccurate predictions. We address this with an explicit memory that preserves per-query history, stabilizing predictions and maintaining track continuity over long videos. Concretely, after extracting frame-level features and initializing query points, we decode the queries using the current frame features $\bh_t$ and the memory $\bM_{t-1}$ propagated from the previous time step.

\vspace{5pt} \boldparagraph{Memory}
The memory maintains a per-query history of visual embeddings for all $N$ queries. Each query has a slot that holds up to $L$ entries. As time advances, updates follow a FIFO policy: when full, the oldest embedding is evicted to accommodate the newest. This yields a bounded, adaptive temporal context whose size does not grow with video length. This design provides a causal, bounded temporal context that favors recent observations, reducing error accumulation and improving robustness to abrupt changes such as long occlusions or fast motion, while avoiding the state overfitting and limited spatial flexibility commonly associated with recurrent memory~\cite{Chen202ARXIV}.

\begin{figure}[b]
    \centering
    \includegraphics[width=1\linewidth]{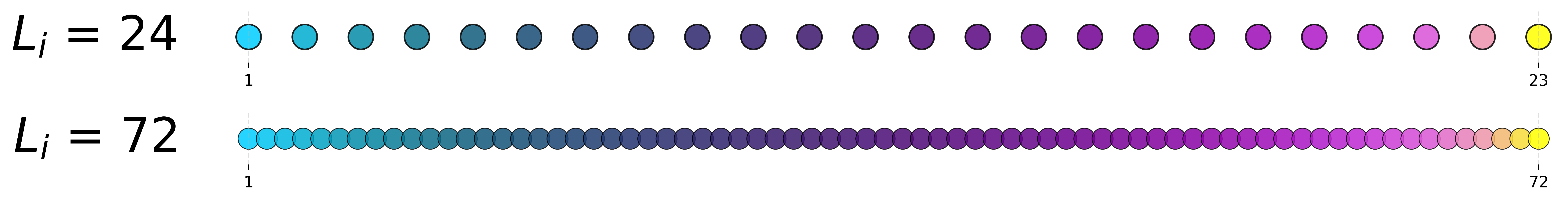}
    \caption{\textbf{Learned temporal embeddings.} We visualize the temporal positional embeddings $\gamma$ by applying PCA to reduce their dimension to 3, then scale each component to $[0,1]$ and map them to RGB. The upper row shows the original embeddings ($L_i=24$), while the lower row shows the embeddings after extension to $L_i=72$.}
    \label{fig:ime_pca}
\end{figure}

\begin{figure}[t]
    \centering
    \includegraphics[width=1\linewidth]{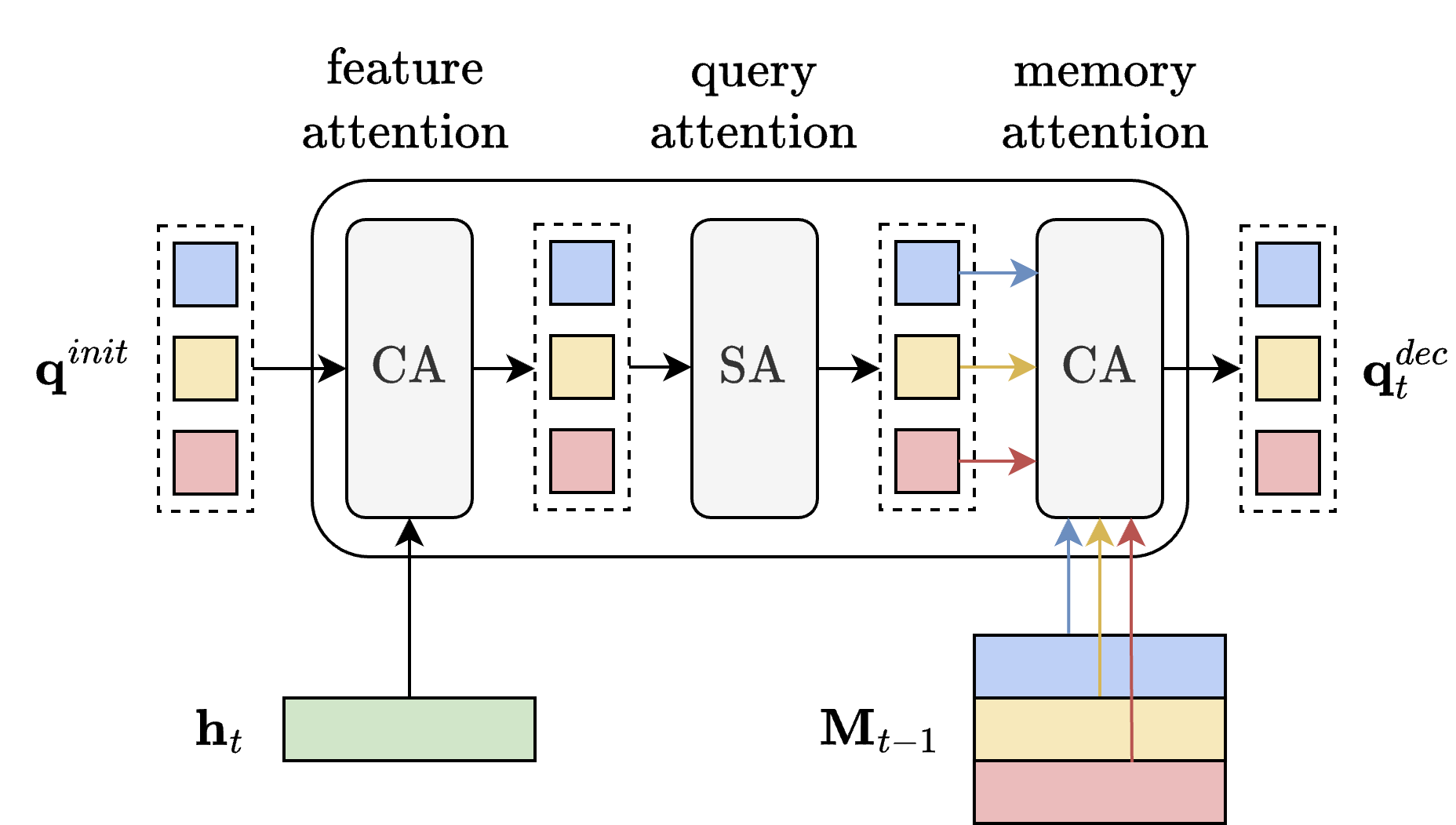}
    \caption{\textbf{Query Decoder.}
    Initialized queries $\bq^{init}$ are updated using current frame features $\bh_t$ and memory $\bM_{t-1}$ from the previous timestep through three attention blocks: feature attention (cross-attention), query attention (self-attention), and memory attention (cross-attention) to each query’s own historical embeddings.
    }
    \label{fig:query_decoder}
\end{figure}

\begin{figure*}[!t]
    \centering
    \subfloat[Top-$k$ Points]{\includegraphics[width=0.4\linewidth]{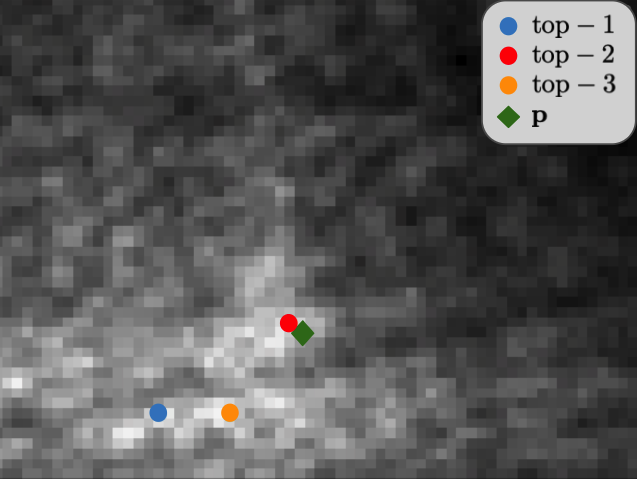}%
    \label{fig:re_ranking:top_k}}
    \hfill
    \subfloat[Re-ranking Module]{\includegraphics[width=0.55\linewidth]{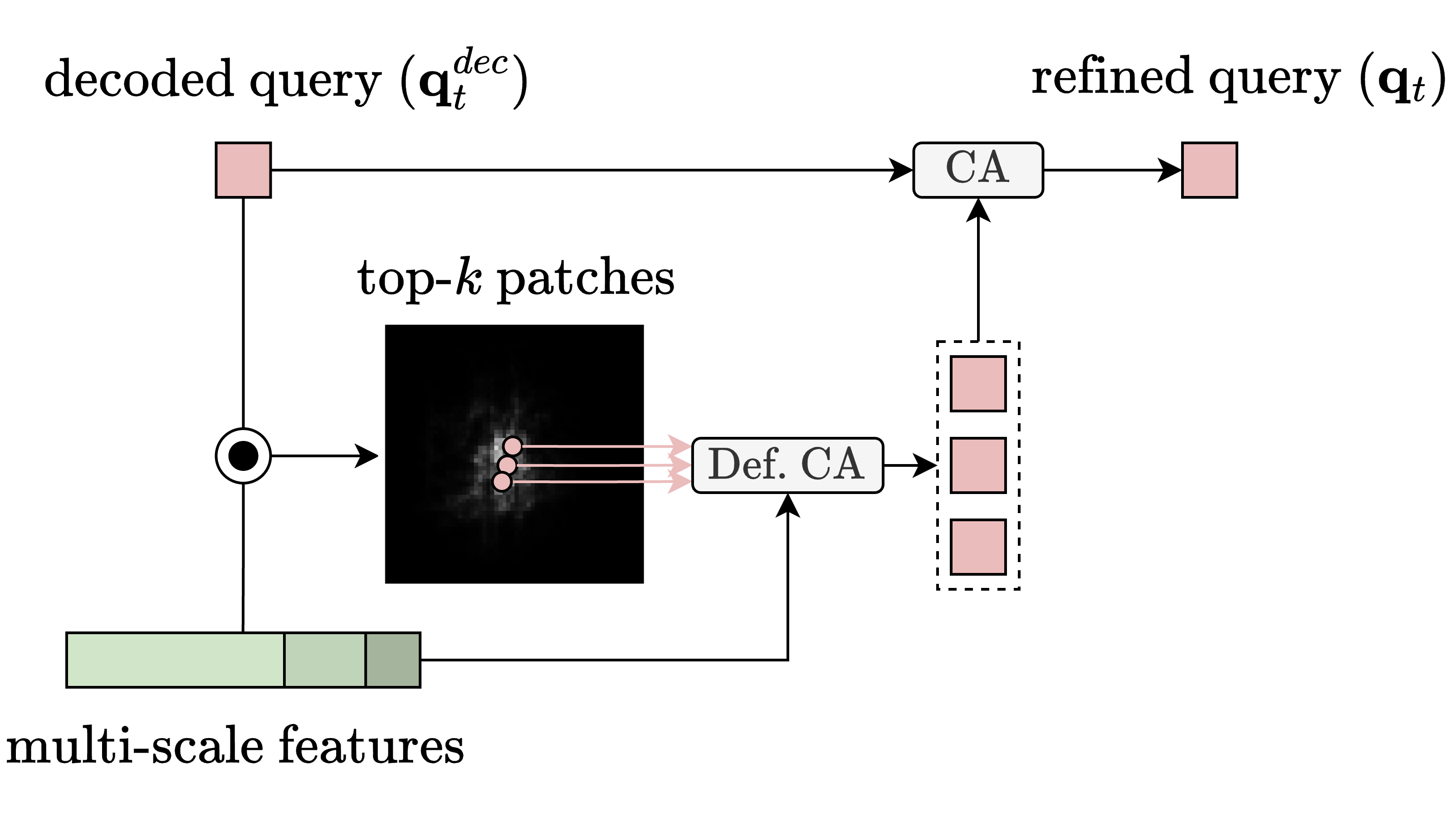}%
    \label{fig:re_ranking:module}}
    \caption{\textbf{Re-ranking.}
    In some cases, a patch with high similarity, though not ranked first, is spatially closer to the ground-truth patch \textbf{(left)}. The top-$3$ patch centers ranked by similarity are shown as dots, while the ground truth is marked by a {\color{Green} diamond}. To address this, we refine decoded queries using a re-ranking module \textbf{(right)}. Starting from $\bq^{dec}_t$ and $\bC^{dec}_t$, we select the top-$k$ patches, extract their multi-scale features via deformable cross-attention (Def.~CA), and fuse them into the queries using cross-attention (CA) to obtain refined queries $\bq_t$ for final similarity prediction.
     }
    \label{fig:re_ranking}
\end{figure*}

\vspace{5pt} \boldparagraph{Decoder} 
We use a standard Transformer decoder with minor adjustments, applied in three sequential attention stages: (i) feature attention, (ii) query self-attention, and (iii) memory attention. Each stage comprises a Multi-Head Attention~(MHA) layer followed by a Feed-Forward Network~(FFN), as shown in \figref{fig:query_decoder}.

In feature attention, initialized queries $\bq^{init}$ attend to current frame features $\bh_t$, aligning with potential matches for similarity-based prediction. In query self-attention, queries interact to promote joint tracking, following CoTracker’s design. In memory attention, each query attends to its own temporal history in $\bM_{t-1}$, retrieving context that improves robustness to drift and occlusion. This combination of current-frame evidence, peer interaction, and temporal memory enables fully online operation with long-term consistency.

Formally, the decoding process is represented as:
\begin{equation}
\bq^{dec}_t = \Phi_{\text{q-dec}} \left( \bq^{init},~ \bh_{t},~ \bM_{t-1} + \gamma \right) \in \mathbb{R}^{N \times D}
\end{equation}
where $\gamma \in \mathbb{R}^{L \times D}$ denotes the learnable temporal positional embeddings added to memory entries, allowing the model to distinguish between closer and more distant history.

\vspace{3pt} \boldparagraph{Inference-time Memory Extension (IME)}
Training uses a fixed memory size $L$, limiting the temporal span the model can attend to. At inference, videos are often much longer. To generalize without retraining, we extend the memory to $L_i > L$ by linearly interpolating the temporal positional embeddings $\gamma$. As memory slots are queried uniformly via attention, this interpolation lets the model attend to longer histories without changing any learned weights. The key is that linear interpolation preserves relative positions among slots, so attention queries remain consistent with training.

To probe what $\gamma$ encodes, we visualize it in \figref{fig:ime_pca} by projecting to 3-dim with PCA and mapping to RGB. 
The top row shows the original embeddings ($L_i=24$); the bottom row shows the extended version ($L_i=72$). The smooth progression from early to recent slots indicates that $\gamma$ encodes temporal order, a structure that is naturally preserved under interpolation.

\vspace{3pt} \boldparagraph{Differences from Track-On}
Our main architectural change from Track-On lies in the memory design. Track-On used two separate modules—spatial and context memories. Track-On2 replaces them with a single, unified memory that stores a per-query temporal history of embeddings. This simplification reduces redundancy and improves efficiency.

\subsubsection{Point Prediction}
\label{sec:point_prediction}

Unlike previous work that regresses the exact location of the points, we formulate the tracking as first a matching problem to one of the patches, that provides a coarse estimate of the correspondence. 
For exact correspondence with higher precision, we further predict offsets to the patch center. Additionally, we also infer the visibility $\hat{\bv}_t \in [0, 1]^{N}$ and uncertainty $\hat{\bu}_t \in [0, 1]^{N}$ for the points of interest.

\vspace{3pt}\boldparagraph{Patch Classification} 
We compute cosine similarity between decoded queries and multi-scale features $\bff^l_t$. The per-scale similarity maps are combined into $\bC^{{dec}}_t$ via a learned weighted average. We then apply a temperature, followed by a spatial softmax over patches in the current frame. The resulting $\bC^{{dec}}_t$ gives, for each query, a similarity distribution over patches.
Training uses a $P$-way classification objective, where the ground-truth class is the patch containing the point of interest ($P$ is the number of patches per frame).

\begin{figure}[h]
    \centering
    \includegraphics[width=0.9\linewidth]{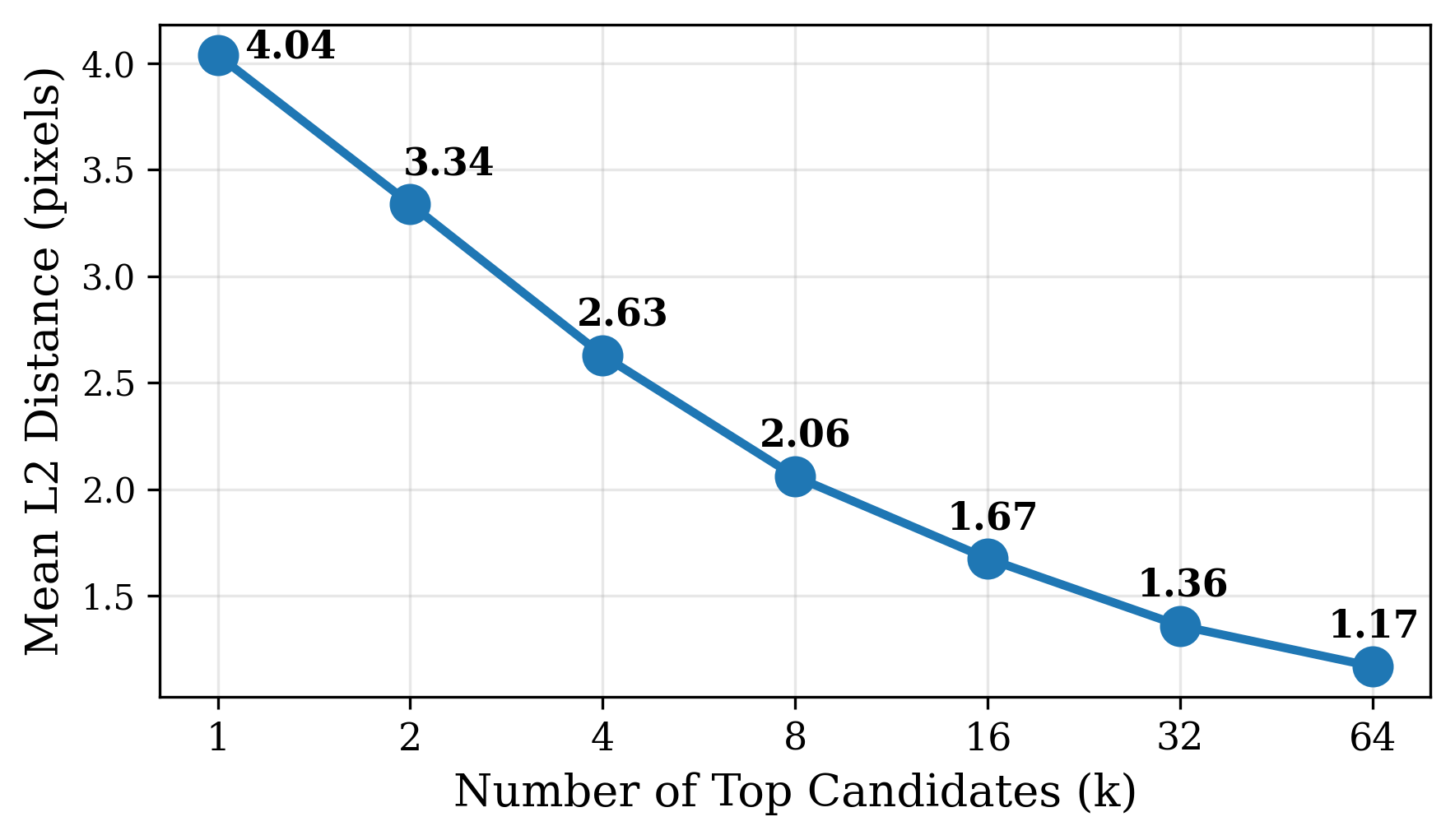}
    \caption{\textbf{Re-ranking oracle.}
    L2 distance to the ground truth versus the number of top-$k$ candidates from the correlation $\bC$, measured by the closest candidate among the top-$k$.}
    \label{rebuttal:fig:reranking_oracle}
\end{figure}

\vspace{3pt}\boldparagraph{Re-ranking}
We observed that the true target patch does not always achieve the highest score in $\bC^{{dec}}_t$, but it typically appears within the top-k. For instance, \figref{fig:re_ranking:top_k} shows a case where the top-2 patch is closer to the true match than the top-1. 
We validate the premise of re-ranking via an oracle analysis on DAVIS in~\figref{rebuttal:fig:reranking_oracle}, measuring the distance from the ground truth to the closest of the top-$k$ predictions in the correlation map $\bC$. As $k$ increases, this distance decreases consistently, indicating that accurate candidates are often present even when the top-1 match is incorrect. Rather than enforcing deterministic selection, using top-$k$ candidates increases the likelihood that the correct match is available for subsequent refinement.
Motivated by this finding, we introduce a re-ranking module $\Phi_{\text{re-rank}}$:
\begin{equation} \label{eq:c_t}
     \bq_t = \Phi_\text{re-rank} \left( \bq^{dec}_t,~ \bff^{\{0, 1, 2, 3\}}_{t},~ \bC^{dec}_t\right) \in \nR^{N \times D}
\end{equation}
where $\bq_t$ denotes the refined queries after ranking.

As illustrated in \figref{fig:re_ranking:module}, we select the top-$k$ patches with the highest decoder similarities and extract their features from the backbone using a deformable cross-attention block (a deformable MHA followed by an FFN). These features are then fused into the original query via a second cross-attention layer (CA), yielding refined queries $\bq_t$. We recompute the similarity map $\bC_t$ from the refined queries and supervise it with a classification loss. The \textbf{center} of the highest-scoring patch is used as the coarse prediction, denoted by $\hat{\bp}^{patch}_t \in \mathbb{R}^{N \times 2}$, which serves as the initial estimate before sub-patch refinement.

To encourage the model to understand local structure around the point of interest, we supervise it with auxiliary uncertainty and confidence signals over the top-$k$ candidates. Specifically, we predict an uncertainty score $\hat{\bu}^{top}_t \in \mathbb{R}^{N \times k}$, where each score indicates whether the top-$k$ candidate is a match, \ie whether the query and candidate correspond to the same point, are visible, and lie within a small spatial error threshold. We also predict a softmax-normalized confidence distribution $\hat{\bs}^{top}_t \in \mathbb{R}^{N \times k}$ over the candidates. These signals are used only during training to guide the local extractor; they are not used in the final prediction.

\vspace{3pt}\boldparagraph{Memory Update}
We append the refined queries $\bq_t$ to the memory. When the memory reaches capacity $L$, we evict the oldest entry (FIFO), maintaining a rolling per-query history. This preserves long-term context without recomputing past representations. As the entries are compact feature vectors, the update is efficient in both compute and storage.

\begin{figure}[t]
    \centering
    \includegraphics[width=1\linewidth]{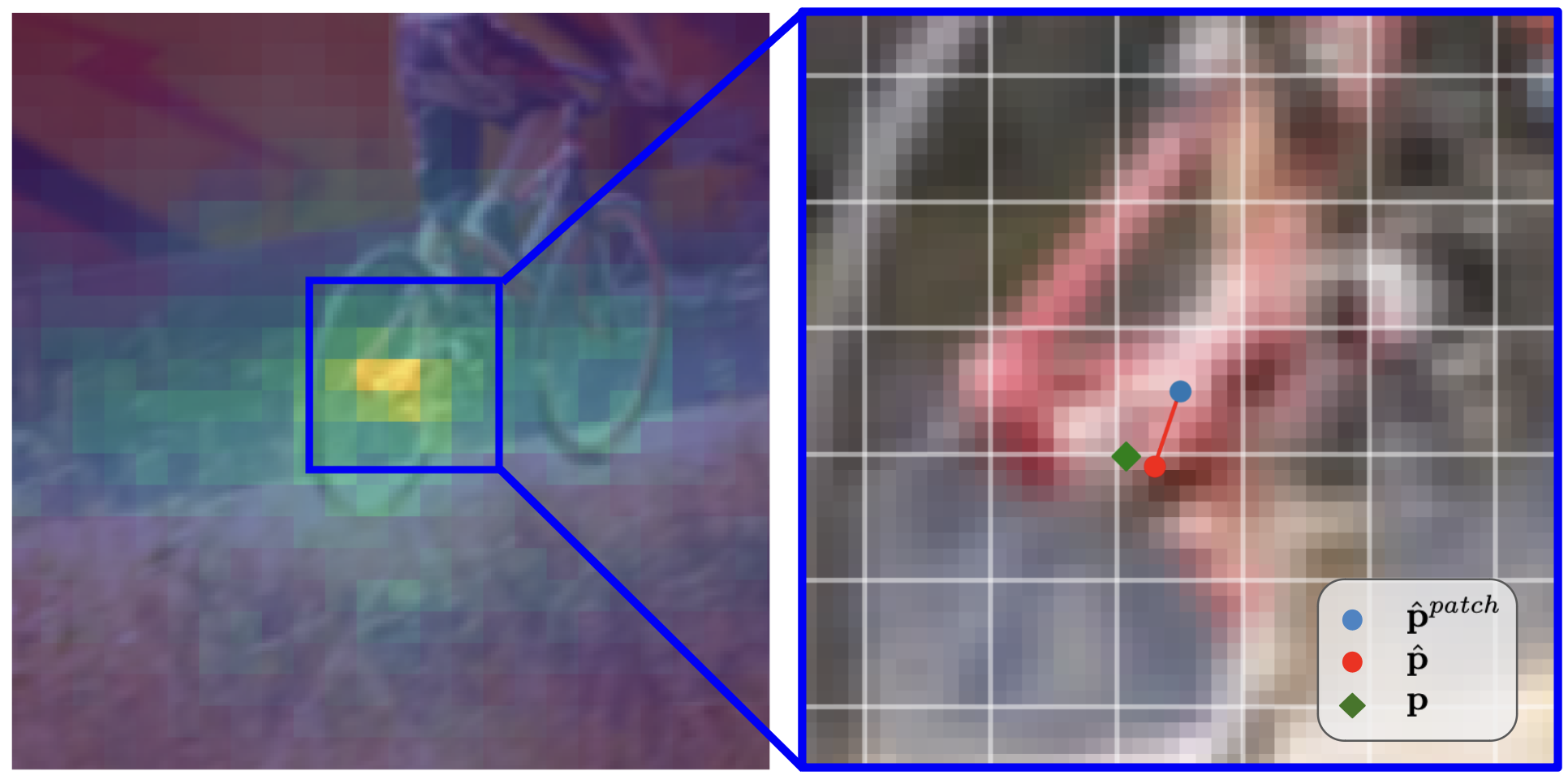}
    \caption{\textbf{Offset Head}. Starting with a rough estimation from patch classification (\textbf{left}), where lighter colors indicate higher correlation, we refine the prediction using the offset head (\textbf{right}). The selected patch center and the final prediction are marked by a {\color{RoyalBlue} blue dot} and a {\color{red} red dot}, respectively, with the ground-truth represented by a {\color{Green} diamond}.}
    \label{fig:offset_head}
\end{figure}

\vspace{3pt}\boldparagraph{Offset Prediction}
For the exact correspondence~($\hat{\bp}_t \in \nR^{N \times 2}$), 
we further predict an offset $\hat{\bo}_t \in \nR^{N \times 2}$ to the patch center by incorporating features from the local region around the inferred patch, as shown in~\figref{fig:offset_head}:
\begin{equation}
\label{eq:patch_refinement}
\hat{\bo}_t = \Phi_{\text{off}}(\bq_t,~\bh_t,~\hat{\bp}^{patch}_t), \quad \quad \quad %
\hat{\bp}_t = \hat{\bp}^{patch}_t + \hat{\bo}_t %
\end{equation}

Here, $\Phi_{\text{off}}$ is a deformable transformer decoder~\cite{Zhu2021ICLR} with 3 layers, excluding self-attention. It refines the query features $\bq_t$ using the multi-scale visual features $\bh_t$, with the reference point set to the coarse prediction $\hat{\bp}^{patch}_t$. This constrains sampling to a local region around the predicted patch. The final offsets are mapped to the range $[-S, S]$ via a $\tanh$ activation, where $S$ is the patch stride. In addition, we estimate the point’s visibility $\hat{\bv}_t$ and uncertainty $\hat{\bu}_t$ via a lightweight MLP applied to the decoded queries $\bq_t$.

At training time, we define a prediction to be uncertain if the prediction error exceeds a threshold ($\delta_u = 12$ pixels) or if the point is occluded. 
During inference, we classify a point as visible if its probability exceeds a threshold $\delta_v$. Although we do not directly utilize uncertainty in our predictions during inference, we found predicting uncertainty to be beneficial for training.

\vspace{3pt} \boldparagraph{Differences from Track-On}
While the overall prediction flow (patch classification followed by offset refinement) remains the same as in Track-On, the use of true multi-scale features introduces key improvements. In Track-On, patch classification was performed on pooled versions of a single-scale feature map. In contrast, Track-On2 directly leverages the multi-scale outputs from the ViT-Adapter. Similarly, offset prediction operates on multi-scale features, rather than approximated scales derived from a single resolution. As a result, Track-On2 is fully multi-scale throughout both patch classification and offset prediction, improving spatial precision and robustness to scale variation. 
In addition, Track-On used a deformable attention head for visibility prediction, which we replace with a single MLP in Track-On2 for simplicity and efficiency.

\subsection{Training}
We train our model using the ground-truth trajectories $\bp_t \in \nR^{N \times 2}$ and visibility information $\bv_t \in \{0,1\}^{N}$. 
For patch classification, we apply cross-entropy loss based on the ground-truth class, patch $\bc^{patch}$.
For offset prediction $\hat{\bo}_t$, we minimize the $\ell_1$ distance between the predicted offset and the actual offset. We supervise the visibility $\hat{\bv}_t$ and uncertainty $\hat{\bu}_t$ using binary cross-entropy loss. 
Additionally, we supervise the uncertainty scores $\hat{\bu}^{top}_t$ and the top-$k$ score distribution $\hat{\bs}^{top}_t$ at the re-ranking stage. Here, we increase the predicted probability of the top-$k$ candidate that is closest to the ground-truth location. The ground-truth score distribution $\bs^{top}_t \in \mathbb{R}^{N \times k}$ is defined as a one-hot vector, where the entry corresponding to the top-$k$ candidate closest to the ground-truth location is set to 1 and all others are set to 0. The total loss is a weighted combination of them:

\begin{equation}
    \label{eq:loss}
    \begin{aligned}
        \cL = &~
        \lambda~ \underbrace{\left(\cL_\text{CE}\left(\bC_t,~\bc^{patch}\right) + 
                  \cL_\text{CE}\left(\bC^{dec}_t,~\bc^{patch}\right)\right)}_{\text{Patch Classification Loss}}  \cdot \bv_t\\
              &~ + \underbrace{\cL_{\ell_1}\left(\hat{\bo}_t,~\bo_t\right)}_{\text{ Offset Loss}} \cdot \bv_t 
              + \underbrace{\cL_\text{BCE}(\hat{\bv}_t, \bv_t)}_{\text{Visibility Loss}} 
              + \underbrace{\cL_\text{BCE}(\hat{\bu}_t, \bu_t)}_{\text{Uncertainty Loss}} \\
              &~ + \underbrace{\cL_\text{BCE}(\hat{\bu}^{top}_t, \bu^{top}_t)}_{\text{Top-$k$ Uncertainty Loss}}
              + \underbrace{\cL_\text{CE}(\hat{\bs}^{top}_t, \bs^{top}_t)}_{\text{Top-$k$ Score Loss}} \cdot \bv_t  
    \end{aligned}
\end{equation}

%% file: sec/04-exp.tex
\section{Experiments}
\label{sec:exp}

\subsection{Experimental Setup}

\boldparagraph{Datasets}  
For training, we use TAP-Vid Kubric~\cite{Doersch2022NeurIPS}, consistent with prior work.
Specifically, we train on the TAP-Vid Kubric variant adopted by CoTracker3~\cite{Karaev2024ARXIV}, a synthetic dataset of roughly 6k video sequences with 120 frames each. We evaluate our model on five datasets with diverse characteristics, covering both real and synthetic domains. Two are subsets of the TAP-Vid benchmark: \textbf{TAP-Vid DAVIS}, consisting of 30 real-world videos from the DAVIS dataset; and \textbf{TAP-Vid Kinetics}, containing over 1{,}000 real-world videos from the Kinetics dataset. Beyond TAP-Vid, we include \textbf{RoboTAP}~\cite{Vecerik2023ICRA}, 265 real-world robotic sequences averaging over 250 frames each; \textbf{Dynamic Replica}~\cite{Karaev2023CVPR}, a 3D reconstruction benchmark with 20 sequences of 300 frames; and \textbf{PointOdyssey (PO)}~\cite{Zheng2023ICCV}, comprising long videos of up to 4{,}325 frames.

\vspace{3pt} \boldparagraph{Metrics}
We evaluate the tracking performance on TAP-Vid subsets and RoboTAP with Occlusion Accuracy (OA), which measures visibility prediction accuracy; \deltaavg, the average proportion of visible points tracked within 1, 2, 4, 8, and 16 pixels; and Average Jaccard (AJ), which jointly assesses visibility and localization precision. For Dynamic Replica, we follow~\cite{Karaev2024ECCV} and report \deltaavg. For PointOdyssey, we adopt two metrics in addition to \deltaavg: Median Trajectory Error (MTE) of visible points; and Survival Rate, defined as the average number of frames as a ratio of video length until failure (threshold is 50 pixels).

\input{tables/sota_A}

\vspace{3pt} \boldparagraph{Evaluation Details}  
We follow the standard protocol of TAP-Vid benchmark by first downsampling the videos to $256 \times 256$. We evaluate models in the queried first protocol, which is the natural setting for causal tracking. In this mode, the first visible point in each trajectory serves as the query, and the task is to track that point in subsequent frames. For datasets with shorter videos~(\eg TAP-Vid DAVIS), we keep the memory size equal to training ($L = L_i = 24$). For longer videos~(\eg TAP-Vid Kinetics, RoboTAP, Dynamic Replica, and PointOdyssey), we extend the memory to $L_i = 72$ to cover a larger temporal span. We add a global support grid of size $20 \times 20$ during inference.

\vspace{3pt} \boldparagraph{Training and Implementation Details}  
We train our model for 200 epochs ($\sim$36K iterations) with a batch size of 32, using the AdamW optimizer~\cite{Loshchilov2019ICLR} on 32$\times$A100 64GB GPUs with mixed precision. The learning rate is set to $5 \times 10^{-4}$ with cosine decay and a linear warmup covering 1\% of training. We use a weight decay of $1 \times 10^{-5}$, gradient clipping at 1.0, and bilinear resizing of input frames to $384 \times 512$. The training-time memory size is set to $L=24$, with top-$k=16$ points. Training samples consist of 48-frame clips from the TAP-Vid Kubric variant adopted by CoTracker3~\cite{Karaev2024ARXIV}, with up to $N=384$ query points. Of these, 75\% are drawn from either the first or middle frame, while the rest are sampled randomly from other frames. We apply random key masking (ratio 0.1) in memory attention during training. Loss coefficients are set with $\lambda=3$, and the offset loss is clipped to the patch stride $S$ to prevent instability from misclassified patches. Deep supervision is applied to the offset head $\Phi_\text{off}$ by averaging losses across layers. The softmax temperature for patch classification is set to $\tau=0.05$, and the visibility threshold to $\delta_v=0.8$. 
Notably, in Track-On, the training video length and memory size were limited to 24 and 12, respectively. In contrast, Track-On2 is trained with longer sequences and expanded memory capacity.

\vspace{-5pt}
\subsection{Quantitative Results}
\label{sec:results}

We compare two variants of our model to prior work: the default model with a DINOv3 ViT-S+ backbone, as described in~\secref{sec:vis_encoder}, and a variant with a DINOv2 ViT-S backbone. 
Comparisons on TAP-Vid DAVIS, TAP-Vid Kinetics, 
and RoboTAP are reported in~\tabref{tab:sota_first}; and on Dynamic Replica and  PointOdyssey in~\tabref{tab:sota_second}. Models are grouped into online and offline categories. 
Offline models, with bidirectional information flow, use either a fixed-size temporal window or the entire video, giving them access to future frames and thus a clear advantage. 
In contrast, online models process one frame at a time, enabling true frame-by-frame inference. We further distinguish models trained or fine-tuned on real-world videos, which generally achieve stronger performance than those trained purely on synthetic TAP-Vid Kubric data, even when using the same architecture. For methods without reported results on certain datasets, we evaluate them using their official public checkpoints and mark these cases accordingly.

\subsubsection{Comparison with Previous Work}

\boldparagraph{TAP-Vid DAVIS}
Track-On2 sets the best results across all metrics. 
While the DINOv3 variant performs strongest, the closest competitor is still our own model with a DINOv2 backbone, differing only by small margins.
Among online methods trained only on synthetic data, it improves over its predecessor Track-On by \textbf{+2.0} AJ, \textbf{+1.9} \deltaavg, and \textbf{+1.2} OA. This boost, specifically on this dataset, is primarily attributed to training with longer video clips, which equips the model with more durable temporal representations (see~\secref{sec:ablation:video_length}). 
It further surpasses strong baselines fine-tuned on real videos: compared to BootsTAPNext-B~\cite{Zholus2025ICCV}, Track-On2 is \textbf{+1.8} AJ (67.0 \vs 65.2), \textbf{+1.4} \deltaavg (79.9 \vs 78.5), and \textbf{+0.8} OA (92.0 \vs 91.2). These results are particularly impressive because our model is an online approach, processing the video frame by frame, and trained only on synthetic data, yet it exceeds the performance of offline models, fine-tuned on real-world data, that process the entire video at once, such as the video variant of CoTracker3~\cite{Karaev2024ARXIV}.

\input{tables/sota_B}

\begin{figure*}[b]
    \centering
    \includegraphics[width=1\linewidth]{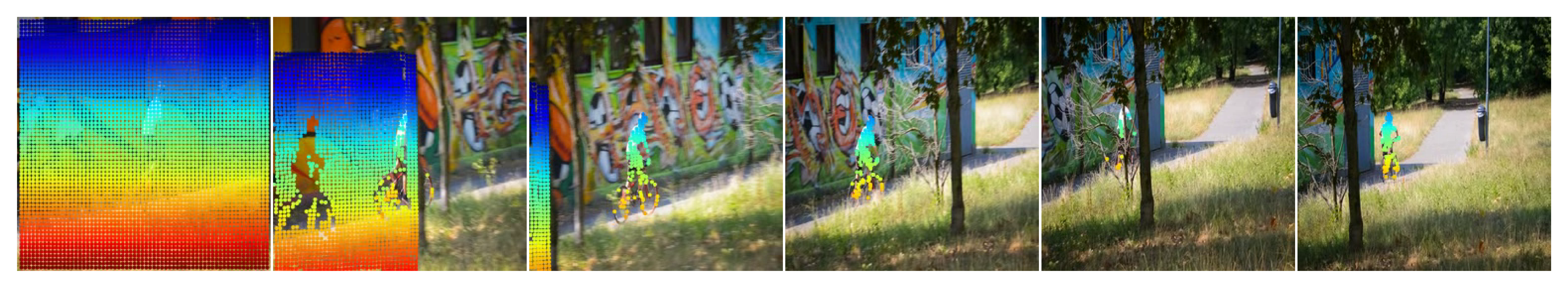}
    \caption{
    \textbf{Dense point tracking on DAVIS.} Uniformly sampled timestamps from a sample DAVIS video, where points are initialized from a $64 \times 64$ uniform grid in the first frame (first column) and tracked across the sequence.
    }
    \label{fig:qual_dense}
\end{figure*}

\vspace{3pt}
\boldparagraph{TAP-Vid Kinetics}
Kinetics comprises long, diverse Internet videos. Within the synthetic-only group, Track-On2 outperforms TAPTRv3~\cite{Qu2024arXiv} in AJ and OA (55.3 \vs 54.5 and 89.6 \vs 88.2) and exceeds TAPNext-B in \deltaavg (69.3 \vs 67.9). 
The difference between our default DINOv3 model and its DINOv2 counterpart remains minor, reinforcing that the gains primarily stem from architectural changes and training setup rather than the backbone alone. 
While real-world fine-tuning improves results across all models on this dataset, with \deltaavg gains of \textbf{+2.8} for TAPNext-B, \textbf{+1.9} for CoTracker3 (Window), and \textbf{+1.3} for CoTracker3 (Video); Track-On2 remains highly competitive without any real-video supervision. It achieves the highest OA and remains close in AJ and \deltaavg to fine-tuned counterparts, despite relying solely on synthetic data.
We also note the parameter efficiency: BootsTAPNext-B uses 194M parameters, whereas our full model has 52.3M parameters with only 23.6M learnable (thanks to frozen visual encoder), \ie roughly one quarter of BootsTAPNext-B’s learnable capacity.

\boldparagraph{RoboTAP}
RoboTAP consists of robotic sequences with an average length of 274 frames, and our model consistently surpasses existing online and offline models across all metrics. 
Among synthetic-only methods, it surpasses TAPTRv3 by \textbf{+3.5} AJ, \textbf{+3.3} \deltaavg, and \textbf{+3.3} OA. This demonstrates that our memory modules, which enable online tracking, are capable of effectively capturing the dynamics of long video sequences despite lacking bidirectional information flow across all frames. 
On this dataset, we also observe a clear benefit from using DINOv3 over DINOv2. While the DINOv2 variant already outperforms previous models in AJ and OA, the DINOv3 backbone further boosts performance. As with Kinetics, real-world fine-tuning brings substantial gains in RoboTAP: AJ increases by \textbf{+5.6} and \textbf{+4.8} for CoTracker3 (Window) and CoTracker3 (Video), respectively. Despite not using any real-video fine-tuning, Track-On2 still outperforms all real fine-tuned baselines by a clear margin. We emphasize that RoboTAP reflects a robotics setting where online operation is crucial; Track-On2’s strong results here indicate its suitability for real-time embodied perception and related tasks.

\boldparagraph{Dynamic Replica}
This dataset comprises 300-frame videos and is suitable to test long-term tracking.
Although CoTracker3 reports the scores of other models on this dataset, their official implementation samples 256 tracks randomly from the available annotations. For fair comparison, we re-evaluated the available checkpoints using uniformly sampled tracks instead of random sampling. On this benchmark, both variants of Track-On2 surpass prior work. The DINOv2 version achieves the highest \deltaavg, outperforming CoTracker3 (Window) by \textbf{+2.3}. The default DINOv3 configuration also outperforms all previous models, but remains slightly behind our DINOv2 variant by 0.1 points. While BootsTAPIR and CoTracker3 remain competitive (68.4 and 72.3 \deltaavg), TAPNext and BootsTAPNext are considerably lower at 47.8 and 46.2. This drop is consistent with the limitation acknowledged in TAPNext, where predictions deteriorate once the sequence length exceeds a threshold, despite strong results on shorter clips. Overall, these results emphasize the importance of temporal robustness and generalization to long sequences; our memory enables stable tracking over hundreds of frames rather than being tied to a fixed clip length.

\begin{figure*}[t]
    \centering
    \includegraphics[width=1\linewidth]{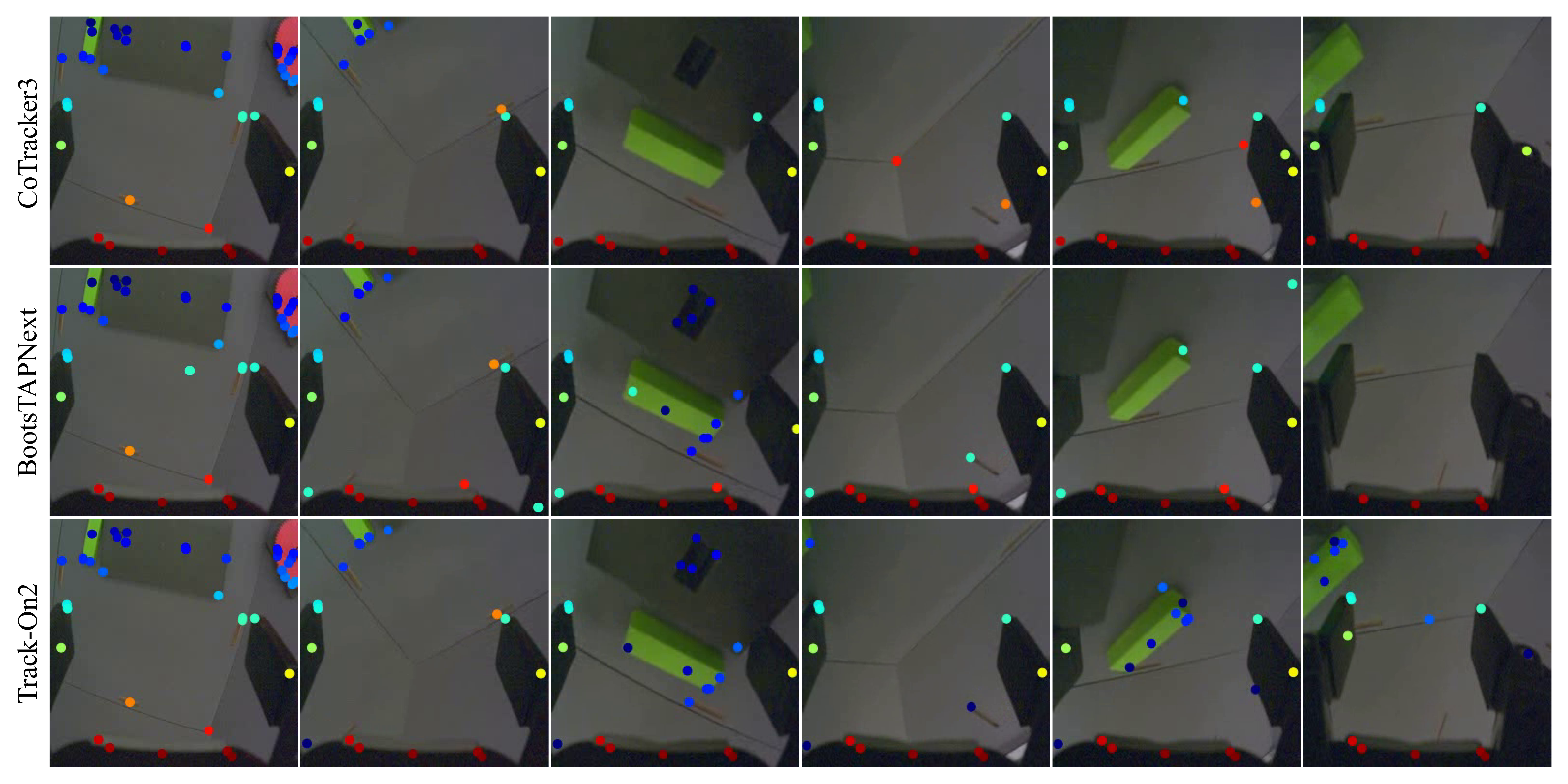}
    \caption{
    \textbf{Sparse point tracking on RoboTAP.} Comparison of CoTracker3 (Window), BootsTAPNext, and Track-On2 on a RoboTAP sequence. Query points are shown in the first column, and subsequent columns show uniformly sampled frames with predictions overlaid. CoTracker3 loses several points on the green block after occlusion and cannot recover, while BootsTAPNext fails to maintain points on the cube later in the sequence. Track-On2 successfully tracks both sets of points throughout the video, demonstrating stronger occlusion recovery and long-term stability.
    }
    \label{fig:qual_comp}
\end{figure*}

\boldparagraph{PointOdyssey}
PointOdyssey is specifically designed to evaluate tracking on very long videos, with an average length of 2,386 frames and a maximum of 4,325. This dataset is particularly important for testing temporal robustness and generalization, since existing models are trained on much shorter sequences: 64 frames for the CoTracker3 family; 48 frames for our model, and the TAPNext family; and 24 frames for Track-On, and BootsTAPIR. This distinction is especially critical for online models, as recovering from errors in fully causal tracking is harder, and mistakes accumulate more severely compared to shorter sequences. Despite this handicap, both variants of Track-On2 achieve the best performance in \deltaavg and survival rate (the fraction of tracks that remain valid without large errors), surpassing even the window-based variant of CoTracker3. Interestingly, the DINOv2 backbone proves more effective than DINOv3 on this dataset: it outperforms DINOv3 across all metrics and even achieves the lowest Median Trajectory Error (MTE) among all models. 

Due to the large number of frames, models that operate at the full-video level, such as BootsTAPIR and the video-input variant of CoTracker3, often run out-of-memory on an NVIDIA A40 (48 GB), since they process all frames simultaneously. CoTracker3 (Window) avoids this by processing at most 16 frames at a time, making it offline but still memory-feasible. In contrast, Track-On and TAPNext are fully online, processing frames one by one. While this avoids heavy memory usage, TAPNext and BootsTAPNext fail on PointOdyssey, as they do in Dynamic Replica, due to the extreme video lengths. As a result, only CoTracker3 (Window) and Track-On2 perform reasonably well on this dataset. Notably, Track-On2 improves upon Track-On by 9.3 points in \deltaavg (a 20\% relative gain) when the same DINOv2 backbone is used, demonstrating stronger temporal modeling and better generalization to long sequences. Crucially, Track-On2 not only avoids failure but also achieves the best results on PointOdyssey, showing its suitability for long-term tracking and its natural fit for streaming scenarios with frame-by-frame processing.

\subsubsection{Comparison to Track-On}
\label{sec:comparison:trackon}

Track-On2 consistently outperforms its predecessor Track-On across all benchmarks. It achieves gains of +2.0 AJ on TAP-Vid DAVIS, +2.3 AJ on Kinetics, and up to +4.6 AJ on RoboTAP, 
when the same DINOv2 backbone is considered. Most notably, the improvement on PointOdyssey is substantial, highlighting Track-On2’s robustness on extremely long sequences. While architectural refinements contribute to these gains, the primary factor is training on longer video clips, which allows the memory module to capture richer temporal context and generalize more effectively to long-term dependencies. 
These contributions are further analyzed in the ablation study (\secref{sec:ablation:trackon}).

\subsubsection{Analysis on Long-term Tracking}
\begin{figure}
    \centering
    \includegraphics[width=1\linewidth]{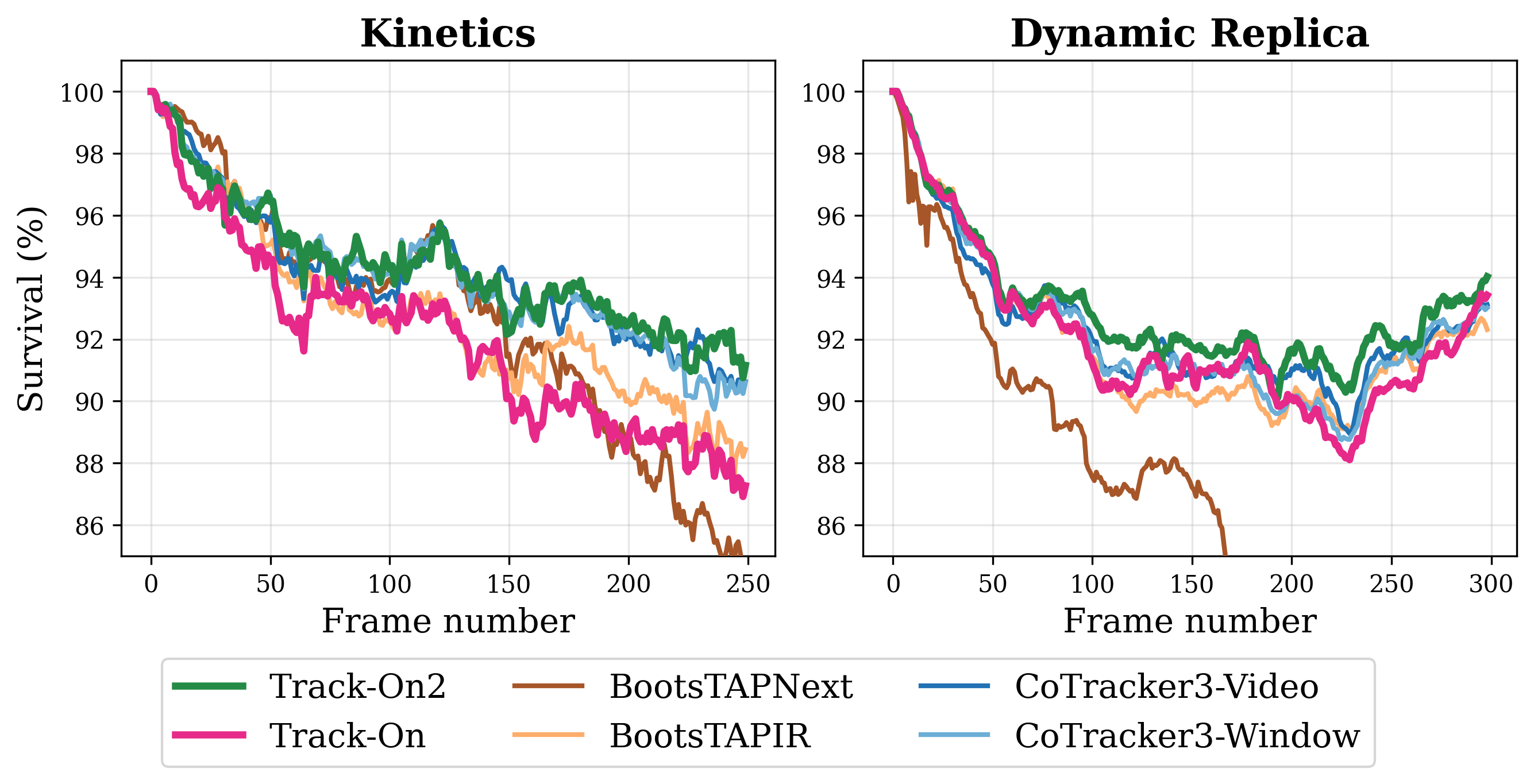}
    \caption{\textbf{Long-term survival comparison.} 
    Survival rate as a function of frame number on Kinetics (left) and Dynamic Replica (right). The survival rate measures the percentage of points that remain successfully tracked over time. }
    \label{rebuttal:fig:survival}
\end{figure}

To further analyze stability without the confounding factor of cumulative failure, we present a frame-wise accuracy analysis in~\figref{rebuttal:fig:survival}. Here, we report the percentage of points with error $<16$px at each timestep, allowing us to visualize how accuracy degrades over time, on Dynamic Replica and Kinetics.
Track-On2 demonstrates the strongest overall robustness. It performs competitively with CoTracker3 on Kinetics; and on Dynamic Replica, our model outperforms all baselines. Notably, while methods like BootsTAPNext suffer sharp drops as sequence length increases, Track-On2 maintains a gentle decay slope, confirming its resistance to long-term drift compared to existing methods, including the original Track-On.

\subsection{Qualitative Results}
\label{sec:qual}

We visualize dense tracking from a $64 \times 64$ grid of query points on a sample video from TAP-Vid DAVIS in~\figref{fig:qual_dense}. The results highlight the model’s ability to maintain trajectories over long spans and to recover after occlusions, with accurate occlusion predictions. 

Additionally, we present sparse tracking results on a RoboTAP video for three top-performing models, namely CoTracker3 (Window), BootsTAPNext, and our model, in~\figref{fig:qual_comp}. Query points are shown on the first frame. We observe that CoTracker3 fails to recover points on the green block once lost (third column), whereas both BootsTAPNext and our model can continue tracking them after occlusion. Later in the sequence, BootsTAPNext loses track of the cube (fifth column), leaving Track-On2 as the only model able to maintain these points until the end. As the video progresses, BootsTAPNext predictions diverge and are predicted occluded, consistent with the limitations reported for TAPNext. Interestingly, after a long static period, CoTracker3 keeps predicting cyan query points on the left robotic arm at their previous location as visible, even though the arm has moved (top row, last two columns). We speculate this is a side effect of its iterative updates, where the model becomes biased towards expected motion rather than appearance cues, assuming the point remains constant after extended inactivity.  
Overall, these results underline the strengths of our design: Track-On2 preserves trajectories over long sequences, recovers from occlusions, and maintains stable predictions where competing methods fail. However, regions with repetitive or textureless appearance remain challenging, as points lack strong anchors for accurate localization.

\begin{figure}[t]
    \centering
    \includegraphics[width=1\linewidth]{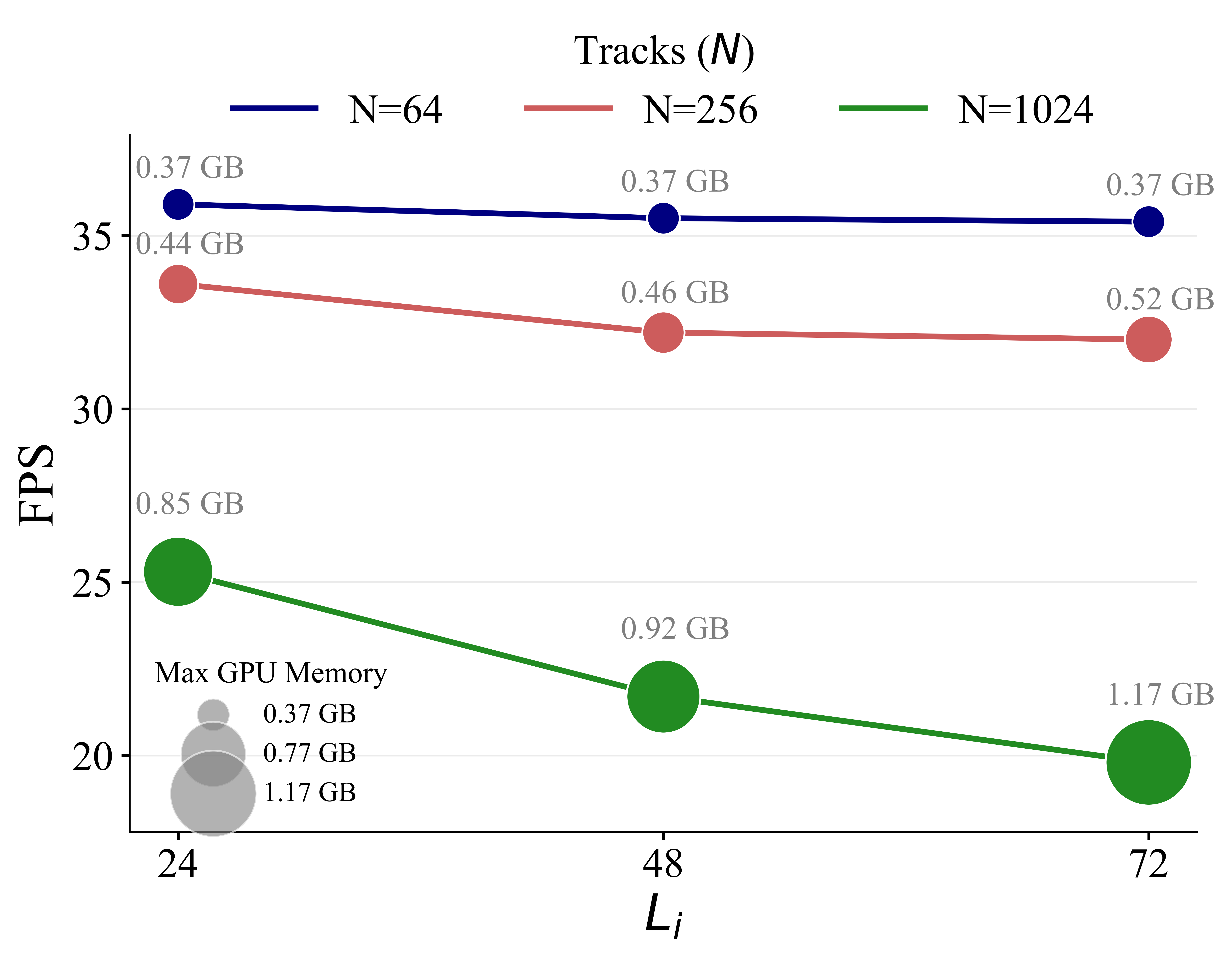}
    \caption{\textbf{Efficiency of tracking with different numbers of tracks ($N$) and memory sizes ($L_i$).} The $x$-axis shows the inference memory size $L_i$, the $y$-axis shows inference speed in frames per second (FPS), and the marker size indicates maximum GPU memory usage (GB). Colors correspond to different numbers of tracks $N$.}
    \label{fig:ablation:efficiency}
\end{figure}

\subsection{Efficiency}
\label{sec:efficiency}

We report inference speed (frames per second, FPS) and peak GPU memory usage as a function of the inference-time memory size $L_i$ for tracking 64, 256, and 1024 points in Fig.~\ref{fig:ablation:efficiency}. All results are obtained using uniform grid queries initialized on the first frame, evaluated in FP32 on a single NVIDIA A100 GPU. 

For sparse tracking with 64 points, our model operates at over 35 FPS while consuming only 0.37\,GB of GPU memory. When tracking 256 points, it maintains close to 35 FPS with less than 0.5\,GB of memory, even as $L_i$ increases. For denser tracking with 1024 points, the peak memory usage remains around 1.2\,GB at $L_i{=}72$, while sustaining approximately 20 FPS. Since state-of-the-art accuracy on longer videos is achieved with higher memory sizes, these results demonstrate that our model meets real-time requirements for practical sparse-point tracking scenarios while maintaining a low memory footprint.

This efficiency stems from our frame-by-frame design, which avoids the large memory footprint of offline models, despite lacking parallel frame processing. In fact, offline methods such as BootsTAPIR or the video-input variant of CoTracker3 often run out-of-memory on long sequences (even with 48 GB GPUs), whereas our model stays below 2.5 GB usage even when tracking over a thousand points. This compact memory footprint also allows Track-On2 to run comfortably on commodity GPUs, making it practical for real-world deployment beyond large-scale servers.

Note that while the predecessor Track-On also operates with a low GPU memory footprint under similar conditions, it achieves lower throughput due to its dual-memory design. In contrast, Track-On2 improves latency significantly, boosting performance from the 20 FPS range to over 30 FPS by unifying the two memory modules into a single, streamlined structure.

\subsection{Ablation Study}
\label{sec:ablation}

On our ablation study, we systematically explore:
\begin{enumerate}[label=(\roman*)]
    \item architectural and training differences between Track-On and Track-On2~(\secref{sec:ablation:trackon}).
    \item choice of DINO backbone variants~(\secref{sec:ablation:dino_backbone});
    \item contributions of model components~(\secref{sec:ablation:components});
    \item training video length and memory size~(\secref{sec:ablation:video_length});
    \item inference-time memory extension~(IME)~(\secref{sec:ablation:ime});
    \item frame sampling strategies~(\secref{sec:ablation:f_sampling});
    \item memory update strategies at inference time~(\secref{sec:ablation:memory_update}).
\end{enumerate}

In particular, we go beyond the limitations of Track-On by increasing the training clip length and systematically exploring how performance is influenced by training video length, memory capacity during training, and inference-time memory size. To better understand how these factors interact under different temporal conditions, we extend our ablations beyond TAP-Vid DAVIS to include RoboTAP, a setting where Track-On did not provide detailed analysis. These experiments offer new insights into the role of training data duration and memory design, a dimension that was largely unexplored in the previous version.

Unless otherwise stated, all models in this section are trained with 48-frame videos, DINOv2 ViT-S as base encoder of ViT Adapter, a memory size of 24, batch size of 32, 384 tracks per batch, 200 epochs ($\sim$36K iterations), and uniform frame sampling with a ratio of 1.0. For RoboTAP experiments, we apply IME with $L_i = 48$. Note that this configuration is not the one reported in our main comparisons, but serves as a baseline to fairly compare models with different settings, as well as the previous version of the model, \ie Track-On, in the extensive ablations.

\subsubsection{Track-On to Track-On2}
\label{sec:ablation:trackon}

\input{tables/exp_trackon_ablation}

We conduct a controlled ablation study in~\tabref{tab:trackon_trackon2_ablation} to analyze the design changes from Track-On to Track-On2. Starting from the original Track-On architecture, we progressively modify one component at a time while keeping all other settings fixed.

We first isolate the effect of the new training regime alone. Comparing the original Track-On baseline with the same architecture trained under the Track-On2 training setup, using the updated dataset and revised frame sampling strategy, shows consistent improvements across datasets, indicating that a substantial portion of the gains arises from improved training rather than architectural modifications.

We next examine the impact of memory design under different training horizons. With short training clips ($T{=}24$), the dual-memory formulation used in Track-On performs better than a unified-memory design~(\secref{sec:vis_encoder}), consistent with prior observations. However, when the training horizon is extended to $T{=}48$, this performance gap largely disappears, suggesting that longer-context training reduces the need for an additional spatial memory. Importantly, increasing the training-time memory size to $L{=}24$ renders the dual-memory formulation infeasible due to out-of-memory errors on 64GB GPUs, whereas the unified-memory design remains stable and scalable. This enables more efficient training and inference, reducing GPU memory usage by approximately $20\%$ and improving inference speed by over 5 FPS when tracking 256 points with $L_i{=}72$.

Finally, we evaluate the remaining architectural refinements. Introducing FPN-style multi-scale feature fusion~(\secref{sec:query_decoder}) and replacing DINOv2 with DINOv3 yields the final Track-On2 model with the strongest overall performance.

Overall, these results show that the primary gains in Track-On2 arise from improved training with longer temporal context, while the architectural choices, particularly unified memory and multi-scale feature fusion, are crucial for making such training feasible, efficient, and scalable. While accuracy gains over Track-On may be modest on short sequences, Track-On2 consistently achieves comparable or improved robustness with substantially better efficiency, especially on long videos.

\subsubsection{DINO as Backbone} 
\input{tables/exp_dino_backbone}
\label{sec:ablation:dino_backbone}

DINOv3 was introduced as a scaled-up version of DINOv2, with larger model capacity and training data, and has shown strong gains across many vision tasks. To study its effect in our framework, we evaluate different backbone options in~\tabref{tab:exp_dino_backbone}. Specifically, we include DINOv2 ViT-S, DINOv3 ViT-S, and DINOv3 ViT-S+. We restrict our study to these smaller variants to maintain efficiency. Since DINOv2 uses a patch size of 14 and DINOv3 uses 16, we upsample the ViT inputs (but not the ViT-Adapter) with a ratio of $16/14$ to match the token count, and apply this upsampling consistently in our main model. Interestingly, despite being trained on a much larger dataset, DINOv3 does not outperform DINOv2 when using the same ViT-S architecture: DINOv2 achieves 0.6 and 0.2 higher AJ scores on DAVIS and RoboTAP, respectively. We attribute this to model capacity, as DINOv3 was primarily designed for large-scale backbones such as ViT-7B, while smaller variants are distilled from it. However, moving from ViT-S (21M parameters) to ViT-S+ (29M, with SwiGLU feedforward layers instead of standard MLPs) yields consistent gains, surpassing DINOv2 on RoboTAP while remaining on par on DAVIS.  

When considering the main comparison tables in~\secref{sec:results}, we find that the DINOv2 variant tends to perform better on synthetic datasets (see~\tabref{tab:sota_second}), whereas DINOv3 provides stronger results on real-world datasets (see~\tabref{tab:sota_first}). Overall, DINOv3 offers only marginal improvements over DINOv2 in our setting, with benefits mostly visible on real-world benchmarks, despite its substantially larger training data.

\subsubsection{Components} 
\label{sec:ablation:components}
\input{tables/exp_components}

We conducted an experiment to examine the impact of each main component, by removing them one at a time while keeping other modules unchanged. 
First, we removed the memory $\bM$. Second, we removed the re-ranking module $\Phi_\text{re-rank}$. Lastly, we removed the offset head $\Phi_\text{off}$, eliminating the calculation of additional offsets. Instead, we used the coarse prediction, \ie the selected patch center, as the final prediction. Note that, we do not apply inference-time memory extension to models in this comparison. From the results in~\tabref{tab:exp_components}, we can make the following observations: 
(i) {Memory is critical}; without it, the model operates only at the frame level and loses all temporal context, leading to AJ drops of 10.0 and 4.7 points on DAVIS and RoboTAP, respectively, and reducing OA on DAVIS from 91.5 to 81.5. (ii) {The re-ranking module} improves all metrics, notably increasing AJ by 4.7 on DAVIS and 2.7 on RoboTAP, as it introduces specialized queries that refine correspondences. (iii) {The offset head} is crucial for fine-grained localization, with \deltaavg dropping by $\sim$6 points on DAVIS and 5.6 on RoboTAP when removed. Without offsets, the model defaults to predicting patch centers; however, unlike memory or re-ranking, this does not directly affect the quality of the queries, but controls the granularity of localization.
Interestingly, OA slightly improves in this case, likely because removing the offset loss term in Eq.~\eqref{eq:loss} rebalances optimization, placing more emphasis on occlusion prediction.

\subsubsection{Video Length and Memory Size}
\label{sec:ablation:video_length}

The number of frames in a training sample is an important design choice, as longer training videos provide supervision over extended temporal spans. In our setting, this directly influences the effective memory size, since memory is the only component that propagates temporal information over time. We evaluate different combinations of training video length $T$ and training memory size $L$, without applying inference-time memory extension (\ie $L = L_i$, see~\secref{sec:ablation:ime}), on TAP-Vid DAVIS, RoboTAP, and PointOdyssey (PO). We report $\delta_{\text{avg}}$ and OA for the first two datasets, and the Survival metric for PO. The Survival metric is particularly relevant for PO’s very long videos, directly assessing robustness in long sequences.

Results in~\tabref{tab:video_memory_length} show that training clip length is the dominant factor, while memory capacity provide complementary gains. Increasing $T$ while keeping memory size fixed (models i $\rightarrow$ ii $\rightarrow$ iii, with $T=24 \rightarrow 36 \rightarrow 48$ and $L=18$) consistently improves performance across all datasets and metrics, with especially large gains in PO Survival. For example, OA on RoboTAP improves by 3.2 points when increasing $T$ from 24 to 48. By contrast, increasing memory size from $L=18 \rightarrow 24$ (model iv) yields modest improvements in DAVIS but slightly reduces performance in RoboTAP, while PO remains unchanged, suggesting that memory capacity alone is less critical than providing sufficiently long training sequences. Finally, we note that the full potential of memory is unlocked when it is extended at inference time, which we discuss in the next section.

\input{tables/exp_video_memory_length}

\subsubsection{Inference-time Memory Extension (IME)} 
\label{sec:ablation:ime}

\begin{figure}[t]
    \centering
    \includegraphics[width=1\linewidth]{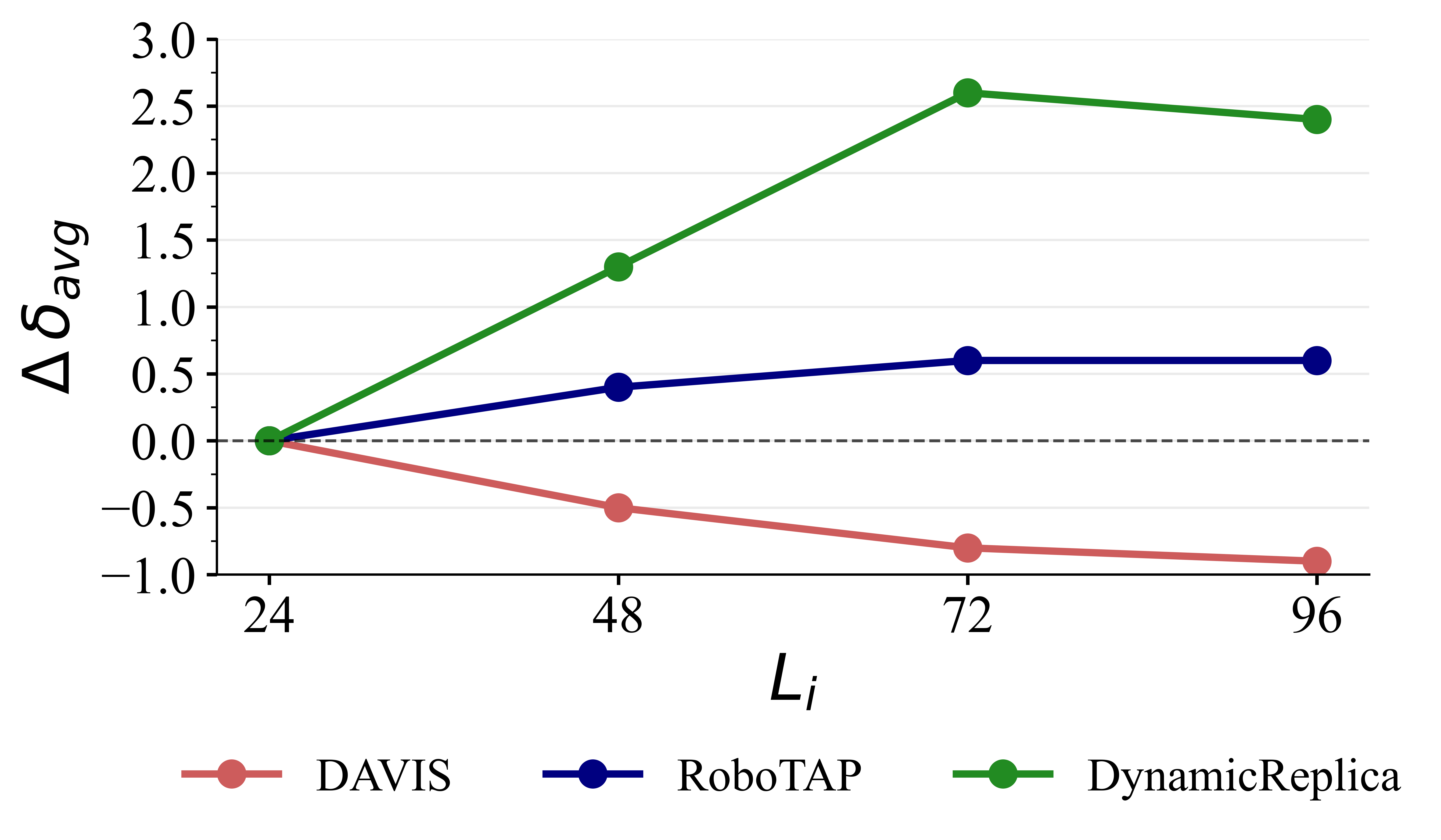}
    \caption{
    \textbf{Impact of inference-time memory size ($L_i$) on performance.} Results are shown for models trained with $L=24$ and $T=48$, reporting the change in \deltaavg when extending memory size to $L_i > L$ on DAVIS, RoboTAP, and Dynamic Replica.
    }
    \label{fig:ablation:ime}
\end{figure}

While models are trained with a fixed memory size $L$, we interpolate them to larger values $L_i$ at inference to better match the characteristics of test videos. We refer to this as inference-time memory extension (IME), as described in~\secref{sec:query_decoder}. To analyze its effect, we apply IME to models trained with $T=48$ and $L=24$, and evaluate the change in $\delta_{avg}$ between the default setting ($L_i = L$) and extended memory sizes ($L_i > L$) on DAVIS, RoboTAP, and Dynamic Replica.
sou
As shown in~\figref{fig:ablation:ime}, performance improves on RoboTAP and Dynamic Replica with increasing $L_i$, while gains saturate or slightly decline beyond a certain length. In contrast, DAVIS shows a steady performance drop as $L_i$ grows. This behavior closely correlates with sequence lengths: DAVIS contains relatively short clips (67 frames on average), while RoboTAP and Dynamic Replica feature significantly longer videos (274 and 300 frames, respectively). In shorter videos like DAVIS, adding more memory entries may introduce noise, as the model is exposed to redundant or rapidly changing visual content. In longer sequences, however, larger memory windows help maintain temporal consistency and prevent drift over time.
The importance of this capability is underscored by recent trends in video understanding, where benchmarks and tasks increasingly require models to operate over hundreds of frames or more~\cite{Wu2024NeurIPS, Weng2024ECCV}. As such, IME becomes a critical component for online point trackers, allowing them to extend beyond their training-time temporal horizon and remain robust in long-form videos without retraining.

\subsubsection{Training Video Frame Sampling}
\label{sec:ablation:f_sampling}

While we sample $T$ frames from the 120-frame videos in the dataset, the way these frames are selected can be important. For example, if the object or camera motion is slow, sampling only sequential frames (stride 1) may provide little useful signal for learning motion. To explore the optimal frame sampling strategy, we experiment with two sampling methods and different ratios, as shown in~\tabref{tab:exp_frame_sampling}.   

Given a training video length $T$ and a ratio $r$, we first take a temporal crop of length $T .r$ (\ie $T . r$ sequential frames). We then either:  
(i) sample frames uniformly, meaning the frame stride is $r$; or  
(ii) randomly sample $T$ frames from the crop, introducing stochastic frame strides. On DAVIS, uniform sampling consistently outperforms random sampling, with a stride of 2 ($r=2.0$) giving the best results, improving AJ by $0.6$ points over stride 1. On RoboTAP, models trained with random sampling perform better than those with uniform sampling, particularly in \deltaavg. Overall, uniform sampling with $r=2.0$ works best for DAVIS, while random sampling with $r=2.0$ is optimal for RoboTAP. However, the uniform $r=2.0$ setting yields the highest average performance (66.9 \vs 66.6 AJ). This suggests that uniform sampling with a stride of 2 provides a good global trade-off, likely by forcing the model to focus on faster-moving objects and thus learning motion cues more effectively. %

Note that the maximum possible value of $r$ while keeping the clip length fixed at $T=48$ is $r = 120 / 48 = 2.5$, given the 120-frame limit of the training videos. However, setting $r$ too high severely limits training diversity, since the starting frame of each clip becomes nearly fixed. By using $r=2.0$, we retain a meaningful stride while still allowing for $120 - 48 \cdot r$ possible offsets per video, which increases data variation. For this reason, we limit our experiments to a maximum stride of $r = 2.0$.

\subsubsection{Memory Update Strategies}
\label{sec:ablation:memory_update}

While Track-On2 adopts a simple FIFO-style memory update during training, memory behavior at inference time can be more flexible, especially for long videos. 
The memory update strategy during training is constrained by the limited temporal span of the training clips, which are restricted to at most 48 frames due to GPU memory constraints. As a result, the memory design must remain simple and well-defined within this regime while still enabling reliable temporal learning. Since we cannot explicitly train on very long-term dependencies under these constraints, we adopt a straightforward FIFO update that refreshes the memory at every frame ($k_1 = 1$).

\input{tables/exp_frame_stride}

However, inference on substantially longer videos allows us to explore alternative update strategies without retraining, provided the temporal ordering of memory entries is preserved. We evaluate different memory update mechanisms in~\tabref{rebuttal:tab:memory_write} on DAVIS, the longer Dynamic Replica (DR), and the very long Point Odyssey (PO). We divide the fixed total memory length $L$ into two partitions, $L_1$ and $L_2$, updated every $k_1$ and $k_2$ frames respectively ($k_2 > k_1$). Here, the $L_2$ partition acts as a long-term memory component, extending the temporal receptive field to approximately $L_1 \cdot k_1 + L_2 \cdot k_2$ frames.
\input{tables/exp_memory_write}

The first two rows of~\tabref{rebuttal:tab:memory_write} use a single partition ($L_2=0$). Our default setting (row i), where $L_1=L$ and $k_1=1$, performs best on DAVIS, where fine-grained temporal propagation is critical for short-term accuracy. Conversely, on longer sequences, incorporating a long-term component consistently improves performance. Experiment (vi), \ie, updating $2L/3$ of the memory every frame and the remaining $L/3$ every 4 frames, improves survival on Point Odyssey by 1.9 points and \deltaavg on Dynamic Replica by 0.4 points. Furthermore, increasing the long-term period from $k_2=2$ to $k_2=4$ yields consistent gains across different splits (comparing (iii) vs. (iv) and (v) vs. (vi)).

Overall, these results suggest that while dense, short-term memory updates are preferable for short videos, incorporating a sparse long-term memory component is beneficial for long-horizon tracking. This analysis highlights that memory in Track-On2 functions as a learned temporal mechanism whose effective behavior depends not only on its size but also on how it is updated over time.

\begin{figure}[t]
    \centering
    \includegraphics[width=1\linewidth]{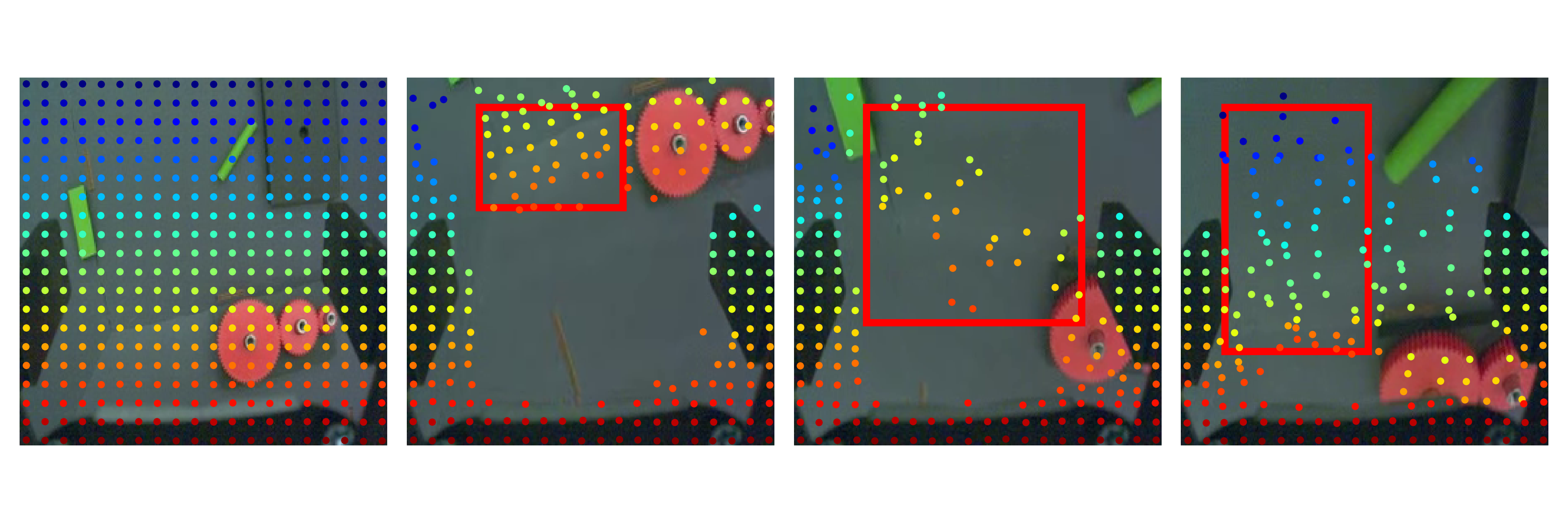}
    \caption{\ga{
    \textbf{Failure cases on uniform regions.} On textureless or repetitive areas (highlighted regions), the spatial structure of uniformly initialized query points degrades over time due to ambiguous appearance-based matching. In contrast, points located on distinctive textured regions (\eg, the red discs) remain stable.
    }}
    \label{fig:failure_case}
\end{figure}

\subsection{Failure Cases}

\ga{
Track-On2 primarily relies on appearance similarity for patch classification, while temporal information is incorporated implicitly through the memory mechanism. Unlike iterative update-based trackers, no explicit spatial location refinement or direct coordinate propagation is enforced across time steps. Although this design improves robustness to error accumulation and enables recovery from drift, it may introduce spatial instability under certain conditions.

In particular, predictions can exhibit jitter or drift in regions with uniform or repetitive texture, where appearance-based matching becomes inherently ambiguous. As illustrated in~\figref{fig:failure_case}, we visualize uniformly initialized grid points on a RoboTAP sequence, sampled every 30 frames. While points located on distinctive textured regions (the red discs) remain stable, the spatial structure of points on textureless regions (highlighted in red rectangles) gradually degrades. This behavior reflects the difficulty of resolving correspondences in areas lacking discriminative visual cues.
Consequently, Track-On2 is most reliable when tracking anchor keypoints with distinctive appearance, whereas uniformly distributed points over textureless regions are more prone to instability.

Another challenging scenario involves thin structures. Due to spatial resolution reduction in intermediate feature maps, fine geometric details may not be sufficiently preserved. As a result, the extracted query feature may not accurately represent the underlying pixel-level structure, leading to imprecise matching along thin surfaces.
}

%% file: tables/sota_A.tex
\begin{table*}[t]
    \centering
    \setlength{\tabcolsep}{6pt}
    \caption{\textbf{Quantitative results on TAP-Vid DAVIS, TAP-Vid Kinetics and RoboTAP.} We compare our model with prior work on TAP-Vid DAVIS, TAP-Vid RGB-Stacking, and RoboTAP under the queried-first setting, reporting AJ, \deltaavg, and OA. We provide results for two variants of Track-On2 that differ only in their frozen backbone: the default DINOv3 and the DINOv2 alternative. Online indicates models that process frames sequentially without access to future frames, enabling frame-by-frame inference. In contrast, offline models can attend to future context, either through full-clip processing or by using a temporal window. Real marks models fine-tuned with real-world videos (highlighted in \raisebox{0pt}[0.9ex][0.4ex]{\colorbox{lightgray}{gray}}), while others are trained purely on synthetic data. \textsuperscript{$\dagger$} indicates results obtained from publicly released checkpoints.} 
    \label{tab:sota_first}

    \begin{tabular}{l!{\vrule width -1pt}c!{\vrule width -1pt}c!{\vrule width -1pt}c!{\vrule width -1pt}c!{\vrule width -1pt}c!{\vrule width -1pt}c!{\vrule width -1pt}c!{\vrule width -1pt}c!{\vrule width -1pt}c!{\vrule width -1pt}c!{\vrule width -1pt}c!{\vrule width -1pt}c!{\vrule width -1pt}c!{\vrule width -1pt}c}
    
        \toprule
        \multicolumn{1}{l}{\multirow{3}{*}{\textbf{Model}}} & \multicolumn{1}{c}{\multirow{3}{*}{Online}} & \multicolumn{1}{c}{\multirow{3}{*}{Real}} & \multicolumn{3}{c}{TAP-Vid DAVIS} & \multicolumn{3}{c}{TAP-Vid Kinetics} & \multicolumn{3}{c}{RoboTAP} \\
        \cmidrule(r){4-6} \cmidrule(r){7-9} \cmidrule(r){10-12} 
         \multicolumn{3}{c}{} &  AJ \up &  \deltaavg \up & \multicolumn{1}{c}{OA \up} & AJ \up  & \deltaavg \up & OA \up & AJ \up  & \deltaavg \up & OA \up \\
         \midrule
         TAPIR & \xmark & \xmark & 56.2 & 70.0 & 86.5 & 49.6 & 64.2 & 85.0 & 59.6 & 73.4 & 87.0 \\
         TAPTR & \xmark & \xmark & 63.0 & 76.1 & 91.1 & 49.0 & 64.4 & 85.2 & 60.1 & 75.3 & 86.9 \\
         TAPTRv2 & \xmark & \xmark & 63.5 & 75.9 & {91.4} & 49.7 & 64.2 & 85.7 & 60.9 & 74.6 & 87.7 \\
         TAPTRv3 & \cmark & \xmark & 63.2 & 76.7 & 91.0 & 54.5 & 67.5 & 88.2  & 64.6 & 77.2 & 90.1 \\
         SpatialTracker & \xmark & \xmark & 61.1 & 76.3 & 89.5 & 50.1 & 65.9 & 86.9 & - & - & - \\ 
         LocoTrack & \xmark & \xmark & 62.9 & 75.3 & 87.2 & 52.9 & 66.8 & 85.3 & 62.3 & 76.2 & 87.1 \\ 
         TAPNext-B & \cmark & \xmark & 62.4 & 76.6 & 90.5 & 53.3 & 67.9 & 87.0 & \repro{59.5} & \repro{72.8} & \repro{88.0} \\
         CoTracker3 (Window) & \xmark & \xmark & 64.5 & 76.7 & 89.7 & 54.1 & 66.6 & 87.1 & 60.8 & 73.7 & 87.1 \\
         CoTracker3 (Video) & \xmark & \xmark & 63.3 & 76.2 & 88.0 & 53.5 & 66.5 & 86.4 & 59.9 & 73.4 & 87.1 \\
         \rowcolor{lightgray}
         BootsTAPIR & \xmark & \cmark & 61.4 & 73.6 & 88.7 & 54.6 & 68.4 & 86.5 & 64.9 & \underline{80.1} & 86.3 \vspace{-1pt} \\
         \rowcolor{lightgray}
         BootsTAPNext-B & \cmark & \cmark & 65.2 & {78.5} & 91.2 & \textbf{57.3} & \textbf{70.6} & 87.4 & \repro{64.0} & \repro{75.0} & \repro{88.7} \vspace{-1pt} \\
         \rowcolor{lightgray}
         CoTracker3 (Window) & \xmark & \cmark & 63.8 & 76.3 & 90.2 & \underline{55.8} & 68.5 & \underline{88.3} & {66.4} & 78.8 & {90.8} \vspace{-1pt} \\
         \rowcolor{lightgray}
         CoTracker3 (Video) & \xmark & \cmark & 64.4 & 76.9 & 91.2 & 54.7 & 67.8 & 87.4 & 64.7 & 78.0 & 89.4 \\
         \midrule
         Track-On  & \cmark & \xmark & 65.0 & 78.0 & 90.8 & 53.9 & 67.3 & 87.8 & 63.5 & 76.4 & 89.4 \\
         Track-On2 (Ours, DINOv2) & \cmark & \xmark & \underline{66.8} & \underline{79.8} & \underline{91.7} & 55.2 & 69.1 & \textbf{89.6} &  \underline{67.1} & 80.0 & \underline{92.7} \\   
         Track-On2 (Ours, DINOv3) & \cmark & \xmark & \textbf{67.0} & \textbf{79.9} & \textbf{92.0} & 55.3 & \underline{69.3} & \textbf{89.6} & \textbf{68.1} & \textbf{80.5} & \textbf{93.4} \\   
        \bottomrule
    \end{tabular}
\end{table*}

%% file: tables/sota_B.tex
\begin{table*}[t]
    \centering
    \caption{\textbf{Quantitative results on Dynamic Replica and PointOdyssey.} 
        We compare our model with prior work under the queried-first setting, reporting results for two variants of Track-On2 that differ only in their frozen backbone: the default DINOv3 and the DINOv2 alternative. Online models process frames sequentially, while offline models use future context (full-clip or temporal window). For methods that exceeded 48 GB GPU memory due to long sequences (PointOdyssey), we report them as out-of-memory. \emph{Real} marks models trained or fine-tuned with real-world videos; others rely only on synthetic data (highlighted in \raisebox{0pt}[0.9ex][0.4ex]{\colorbox{lightgray}{gray}}). All results are obtained using the official publicly released checkpoints.        
        }
    \label{tab:sota_second}

    \begin{tabular}{l!{\vrule width -1pt}c!{\vrule width -1pt}c!{\vrule width -1pt}c!{\vrule width -1pt}c!{\vrule width -1pt}c!{\vrule width -1pt}c}
        \toprule
        \multicolumn{1}{l}{\multirow{3}{*}{\textbf{Model}}} & \multicolumn{1}{c}{\multirow{3}{*}{Online}} & \multicolumn{1}{c}{\multirow{3}{*}{Real}} & \multicolumn{1}{c}{Dynamic Replica} & \multicolumn{3}{c}{PointOdyssey} \\
        \cmidrule(r){4-4} \cmidrule(r){5-7} 
         \multicolumn{3}{c}{} & 
         \multicolumn{1}{c}{\deltaavg \up} & 
         \deltaavg \up & 
         MTE $\downarrow$ &
         \multicolumn{1}{c}{Survival \up} \\
         \midrule
         TAPNext-B & \cmark & \xmark & {47.8} & {9.5} & {86.3} & {16.0} \\

         \rowcolor{lightgray}
         BootsTAPIR & \xmark & \cmark & 68.4 & \multicolumn{3}{c}{Out-of-Memory} \vspace{-1pt}  \hspace{-4.5pt} \\
         \rowcolor{lightgray}
         BootsTAPNext-B & \cmark & \cmark & 46.2 & {9.9} & {88.2} & {12.8} \vspace{-1pt} \\
         \rowcolor{lightgray}
         CoTracker3 (Window) & \xmark & \cmark & 72.3 & {{44.5}} & {\underline{20.7}} & {{56.3}} \vspace{-1pt} \\
         \rowcolor{lightgray}
         CoTracker3 (Video) & \xmark & \cmark & 72.3 & \multicolumn{3}{c}{Out-of-Memory}  \vspace{-1pt} \hspace{-4.5pt} \\
         \midrule
         Track-On  & \cmark & \xmark & 73.2 & 38.1 & 28.8 & 49.5 \\
         Track-On2 (Ours, DINOv2)  & \cmark & \xmark & \textbf{74.6} & \textbf{47.4} & \textbf{20.5} & \textbf{57.8}  \\    
         Track-On2 (Ours, DINOv3) & \cmark & \xmark & \underline{74.5} & \underline{45.1} & {22.0} & \underline{57.7}  \\    
        \bottomrule
    \end{tabular}
\end{table*}

%% file: tables/exp_trackon_ablation.tex
\begin{table}[t]
\centering
\setlength{\tabcolsep}{2.7pt}
\caption{\textbf{Ablation from Track-On to Track-On2.}
Dv3 indicates a DINOv3 backbone; FPN denotes multi-scale feature fusion; Mem.\ denotes a unified single-memory design; Data indicates training with the Track-On2 data and sampling strategy; $T$ and $L$ are the training clip length and training-time memory size. Row (i), shaded in gray, corresponds to the original Track-On trained with the original data regime; all other variants use the Track-On2 training setup.}
\label{tab:trackon_trackon2_ablation}

\begin{tabular}{l!{\vrule width -1pt}c!{\vrule width -1pt}c!{\vrule width -1pt}c!{\vrule width -1pt}c!{\vrule width -1pt}c!{\vrule width -1pt}c | c!{\vrule width -1pt}c!{\vrule width -1pt}c!{\vrule width -1pt}c!{\vrule width -1pt}c}
\toprule
&
\multicolumn{1}{c}{\multirow{3}{*}{Dv3}} &
\multicolumn{1}{c}{\multirow{3}{*}{FPN}} &
\multicolumn{1}{c}{\multirow{3}{*}{Mem.}} &
\multicolumn{1}{c}{\multirow{3}{*}{Data}} &
\multicolumn{1}{c}{\multirow{3}{*}{$T$}} &
\multicolumn{1}{c}{\multirow{3}{*}{$L$}} &
\multicolumn{2}{c}{DAVIS} &
\multicolumn{2}{c}{RoboTAP} &
PO \\
\cmidrule(lr){8-9}\cmidrule(lr){10-11}\cmidrule(lr){12-12}
 & & & & & & &
 $\delta^{x}_{avg}$ & OA &
 $\delta^{x}_{avg}$ & OA &
 Survival \\
\midrule

\rowcolor{lightgray}
(i) & \xmark & \xmark & \xmark & \xmark & 24 & 12
& 78.0 & 90.8 & 76.4 & 89.4 & 49.5 \\

\midrule

(ii) & \xmark & \xmark & \xmark & \cmark & 24 & 12
& 78.9 & 91.3 & 79.4 & 92.8 & 53.9 \\

(iii) & \xmark & \xmark & \xmark & \cmark & 48 & 12
& 79.1 & 91.5 & 80.0 & 92.9 & 57.5 \\

(iv) & \xmark & \xmark & \xmark & \cmark & 48 & 24
& \multicolumn{5}{c}{Training OOM} \\

\midrule

(v) & \xmark & \xmark & \cmark & \cmark & 24 & 12
& 78.8 & 90.5 & 79.3 & 92.1 & 50.8 \\

(vi) & \xmark & \xmark & \cmark & \cmark & 48 & 12
& 78.9 & 91.5 & 80.6 & 92.8 & 54.4 \\

(vii) & \xmark & \xmark & \cmark & \cmark & 48 & 24
& 78.7 & 91.8 & 79.9 & 92.8 & 57.4 \\

(viii) & \xmark & \cmark & \cmark & \cmark & 48 & 24
& 79.8 & 91.7 & 80.0 & 92.7 & 57.8 \\

(ix) & \cmark & \cmark & \cmark & \cmark & 48 & 24
& 79.9 & 92.0 & 80.5 & 93.4 & 57.7 \\

\bottomrule
\end{tabular}
\end{table}

%% file: tables/exp_dino_backbone.tex
\begin{table}[t]
    \centering
        \caption{\textbf{DINO backbones in ViT-Adapter.} We replace the DINOv2 backbone in our ViT-Adapter with DINOv3. All metrics are higher-is-better.}
        \begin{tabular}{lc ccc ccc}
             \toprule
             \multicolumn{1}{l}{\multirow{3}{*}{\textbf{Backbone}}} &
             \multicolumn{1}{c}{\multirow{3}{*}{\textbf{Arch.}}} & 
             \multicolumn{3}{c}{DAVIS} & \multicolumn{3}{c}{RoboTAP} \\
             \cmidrule(r){3-5} \cmidrule(r){6-8}  
             & & AJ & \deltaavg & \multicolumn{1}{c}{OA} & AJ  & \deltaavg & OA \\
             \midrule
             \multirow{1}{*}{DINOv2} & ViT-S & \textbf{66.2} & \textbf{79.1} & \textbf{91.5} & 66.6 & 79.4 & \textbf{92.3} \\
             \midrule
             \multirow{2}{*}{DINOv3} & ViT-S & 65.6 & 78.8 & 90.7 & 66.4 & 79.5 & 92.2 \\
              & ViT-S+ & \textbf{66.2} & 79.0 & 91.1 & 66.9 & \textbf{79.7} & \textbf{92.3} \\
           \bottomrule
        \end{tabular}
        \label{tab:exp_dino_backbone}
    
\end{table}

%% file: tables/exp_components.tex
\begin{table}[t]
    \centering
        \caption{\textbf{Model Components.} Removing individual components of our model, namely memory ($\bM$) the re-ranking module ($\Phi_\text{re-rank}$), and offset head ($\Phi_\text{off}$), one at a time. All metrics are higher-is-better.}
        \begin{tabular}{l ccc ccc}
             \toprule
             \multicolumn{1}{l}{\multirow{3}{*}{\textbf{Model}}} & \multicolumn{3}{c}{DAVIS} & \multicolumn{3}{c}{RoboTAP} \\
            \cmidrule(r){2-4} \cmidrule(r){5-7} 
            &  AJ &  \deltaavg & \multicolumn{1}{c}{OA} & AJ  & \deltaavg & OA \\
             \midrule
             Full Model & \textbf{66.2} & \textbf{79.1} & \textbf{91.5} & \textbf{66.6} & \textbf{79.4} & \textbf{92.3} \\
             - No memory & 56.2 & 70.8 & 81.5 & 61.9 & 75.6 & 88.3 \\
             - No re-ranking & 61.5 & 75.4 & 87.8 & 63.9 & 76.7 & 89.9 \\
             - No offset head & 60.5 & 73.1 & 90.8 & 61.3 & 73.8 & 92.7 \\
           \bottomrule
        \end{tabular}
        \label{tab:exp_components}
    
\end{table}

%% file: tables/exp_video_memory_length.tex
\begin{table}[t]
    \centering
    \caption{\textbf{Training video length and memory size.} We compare different training video lengths $T$ and memory sizes $L$, without applying inference-time memory extension (\ie $L=L_i$). %
    All metrics are higher-is-better.}
    \begin{tabular}{l cc cc cc c}
         \toprule
         \multicolumn{1}{l}{\multirow{3}{*}{}} & 
         \multicolumn{1}{l}{\multirow{3}{*}{$T$}} & 
         \multicolumn{1}{l}{\multirow{3}{*}{$L$}} & 
         \multicolumn{2}{c}{DAVIS} &
         \multicolumn{2}{c}{RoboTAP} &
         \multicolumn{1}{c}{PO} \\
         \cmidrule(r){4-5} \cmidrule(r){6-7} \cmidrule(r){8-8}  
         & & & \deltaavg & OA & \deltaavg & OA & Survival\\
         \midrule
         (i) & 24 & 18 & 77.5 & 90.2 & 76.0 & 89.0 & 46.8 \\
         (ii) & 36 & 18 & 78.3 & 90.7 & 78.2 & 91.2 & 51.3 \\
         (iii) & 48 & 18 & 79.0 & 90.7 & \textbf{79.1} & \textbf{92.2} & \textbf{51.5} \\
         (iv) & 48 & 24 & \textbf{79.1} & \textbf{91.5} & 78.9 & 91.5 & \textbf{51.5} \\
         \bottomrule
    \end{tabular}
    \label{tab:video_memory_length}
\end{table}

%% file: tables/exp_frame_stride.tex
\begin{table}
    \centering
    \setlength{\tabcolsep}{5pt}
    \caption{\textbf{Training video frame sampling.} Comparison of uniform and random strategies with different ratios $r$, where $r$ controls the temporal crop size before selecting $T$ training frames. All metrics are higher-is-better.}
    \begin{tabular}{cc ccc ccc}
        \toprule
        \multicolumn{1}{l}{\multirow{3}{*}{\textbf{Sampling}}} & 
        \multicolumn{1}{c}{\multirow{3}{*}{\textbf{Ratio}}} & 
        \multicolumn{3}{c}{DAVIS} &
        \multicolumn{3}{c}{RoboTAP} \\
        \cmidrule(r){3-5} \cmidrule(r){6-8} 
        & & AJ &  \deltaavg & \multicolumn{1}{c}{OA} & AJ  & \deltaavg & OA \\
        \midrule
        \multirow{2}{*}{Uniform} & 1.0 & 66.2 & 79.1 & 91.5 & 66.6 & 79.4 & 92.3 \\
         & 2.0 & \textbf{66.8} & \textbf{79.8} & \textbf{91.8} & 67.0 & 79.9 & \textbf{92.7} \\
        \midrule
        \multirow{2}{*}{Random} & 1.5 & 65.7 & 78.8 & 90.1 & 67.2 & 80.4 & 92.1 \\
         & 2.0 & 65.9 & 78.9 & 90.9 & \textbf{67.3} & \textbf{80.5} & 92.1 \\
       \bottomrule
    \end{tabular}
    \label{tab:exp_frame_sampling}
\end{table}

%% file: tables/exp_memory_write.tex
\begin{table}[b]
    \centering
    \setlength{\tabcolsep}{4pt}
    \caption{\textbf{Memory update mechanisms.}
    Comparison of different memory update strategies.
    The total memory length $L$ is fixed and split into short-term ($L_1$) and long-term ($L_2$) components, updated every $k_1$ and $k_2$ frames, respectively. Results are reported on DAVIS, Dynamic Replica (DR), and Point Odyssey (PO). All metrics are higher-is-better.
    }
    \begin{tabular}{lcc cc | ccc c cc}
        \toprule
        &
        \multicolumn{2}{c}{\textbf{Ratio}} & 
        \multicolumn{2}{c}{\textbf{Period}} & 
        \multicolumn{3}{c}{DAVIS} &
        \multicolumn{1}{c}{DR} &
        \multicolumn{2}{c}{PO} \\
        \cmidrule(r){2-3} \cmidrule(r){4-5} \cmidrule(r){6-8} \cmidrule(r){9-9} \cmidrule(r){10-11} 
        & $L_1$ & $L_2$ & $k_1$ & $k_2$ & AJ &  \deltaavg & \multicolumn{1}{c}{OA} & \deltaavg & \deltaavg & Survival  \\
        \midrule
        (i) & 1 & - & 1 & - & \textbf{67.0} & \textbf{79.9} & \textbf{92.0} & 74.5 & 45.1 & 57.7 \\
        (ii) & 1 & - & 2 & - & 66.4 & 79.0 & 91.6 & 74.8 & 43.9 & 55.0 \\
        (iii) & $1/2$ & $1/2$ & 1 & 2 & 66.5 & 79.4 & 91.8 & 74.5 & 44.4 & 58.4 \\
        (iv) & $1/2$ & $1/2$ & 1 & 4 & 66.3 & 79.2 & 91.5 & 74.4 & 46.4 & 58.9 \\
        (v) & $2/3$ & $1/3$ & 1 & 2 & 66.6 & 79.6 & 91.8 & 74.8 & 45.8 & 58.5 \\
        (vi) & $2/3$ & $1/3$ & 1 & 4 & 66.3 & 79.2 & 91.6 & \textbf{74.9} & \textbf{48.5} & \textbf{59.6} \\
       \bottomrule
    \end{tabular}
    \label{rebuttal:tab:memory_write}
\end{table}

%% file: sec/05-conclusion.tex
\section{Conclusion}

In this work, we presented Track-On2, a simplified yet more powerful model for online long-term point tracking. By consolidating and refining the architecture of its predecessor, Track-On, we arrived at a design that is both faster and more accurate, enabling efficient frame-by-frame inference even on very long sequences. Our analysis demonstrated that memory plays a central role in temporal reasoning, and we systematically studied how training video length, frame sampling strategy, and inference-time memory extension affect overall performance. These insights provide practical guidance for developing causal trackers under resource constraints. Through extensive experiments, we showed that Track-On2 consistently outperforms prior methods, including offline models with access to future frames and real-data fine-tuned baselines, despite being trained exclusively on synthetic videos. The ability to generalize from synthetic supervision alone underscores the value of principled training design and highlights memory-based causal models as a strong foundation for scalable, real-world tracking. Looking forward, we believe Track-On2 opens new opportunities for broader video understanding tasks, ranging from dense 3D reconstruction to robotics and embodied perception.

\vspace{3pt} \boldparagraph{Limitations and Future Work}
\ga{
While several design choices in Track-On2 naturally improve efficiency (\eg, strict frame-by-frame processing, a single expandable memory, and the absence of heavy 3D cost volumes), the model is not explicitly engineered for hardware-specific optimization. In particular, we rely on standard Transformer operations without tailoring low-level implementations for reduced FLOPs or memory footprint (\eg, attention kernel redesign, tensor operation fusion, or hardware-aware concatenation strategies), unlike recent object tracking works that optimize both architectural design and low-level operations for edge deployment~\cite{Borsuk2022ECCV, Blatter2023WACV, Kang2023ICCV}. 
As a result, although Track-On2 achieves competitive runtime on high-end GPUs (all benchmarks are conducted on NVIDIA A100), it has not yet been optimized for resource-constrained environments such as CPUs or embedded devices. Extending Track-On2 with hardware-aware optimization and system-level efficiency improvements is therefore a valuable direction for future work, particularly given the practical importance of real-time tracking in embedded and real-world applications.

Beyond efficiency, the failure cases discussed above highlight the limitations of relying predominantly on appearance-based matching. 
}
Explicitly modeling motion within the architecture is a promising direction for future work, as it could improve temporal coherence while reducing over-reliance on appearance. Future work also includes exploring more advanced memory update strategies beyond FIFO, such as adaptive memory writes conditioned on video length or query content, to better balance short-term and long-term information in very long sequences. More broadly, effective adaptation to real-world data remains an important challenge for point tracking models. Exploring real-world fine-tuning and improving pseudo-labeling strategies are key directions for enhancing robustness and generalization beyond synthetic training regimes.

%% file: sec/07-ack.tex
\vspace{-3pt}
\section{Acknowledgements}
This project is funded by the European Union (ERC, ENSURE, 101116486) with additional compute support from Leonardo Booster (EuroHPC Joint Undertaking, EHPC-AI-2024A05-028). Views and opinions expressed are however those of the author(s) only and do not necessarily reflect those of the European Union or the European Research Council. Neither the European Union nor the granting authority can be held responsible for them. Weidi Xie would like to acknowledge 
the Scientific Research Innovation Capability Support Project for Young Faculty (ZYGXQNJSKYCXNLZCXM-I22).

%% file: sec/06-bio.tex
\begin{IEEEbiography}[{\includegraphics[width=1in,height=1.25in,clip,keepaspectratio]{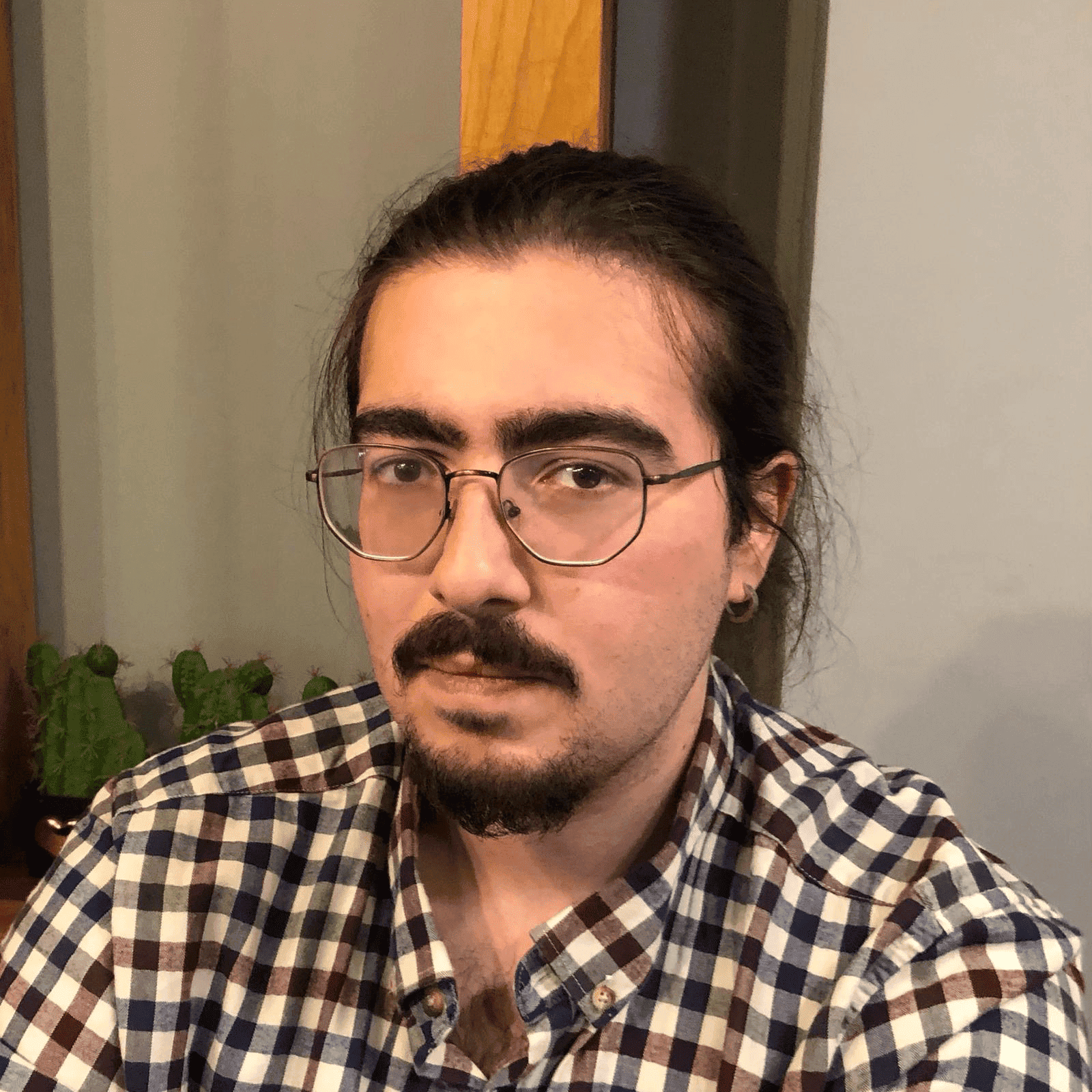}}]{Görkay Aydemir} received his B.Sc. degree in computer engineering from Middle East Technical University, Ankara, Turkey, in 2022, and his M.Sc. degree in computer science and engineering from Koç University, Istanbul, Turkey, in 2025. His research interests include point tracking, motion estimation, and object-centric learning. He regularly serves as reviewer at leading CV and ML conferences, including ICLR, NeurIPS, and CVPR.
\end{IEEEbiography}

\begin{IEEEbiography}[{\includegraphics[width=1in,height=1.25in,clip,keepaspectratio]{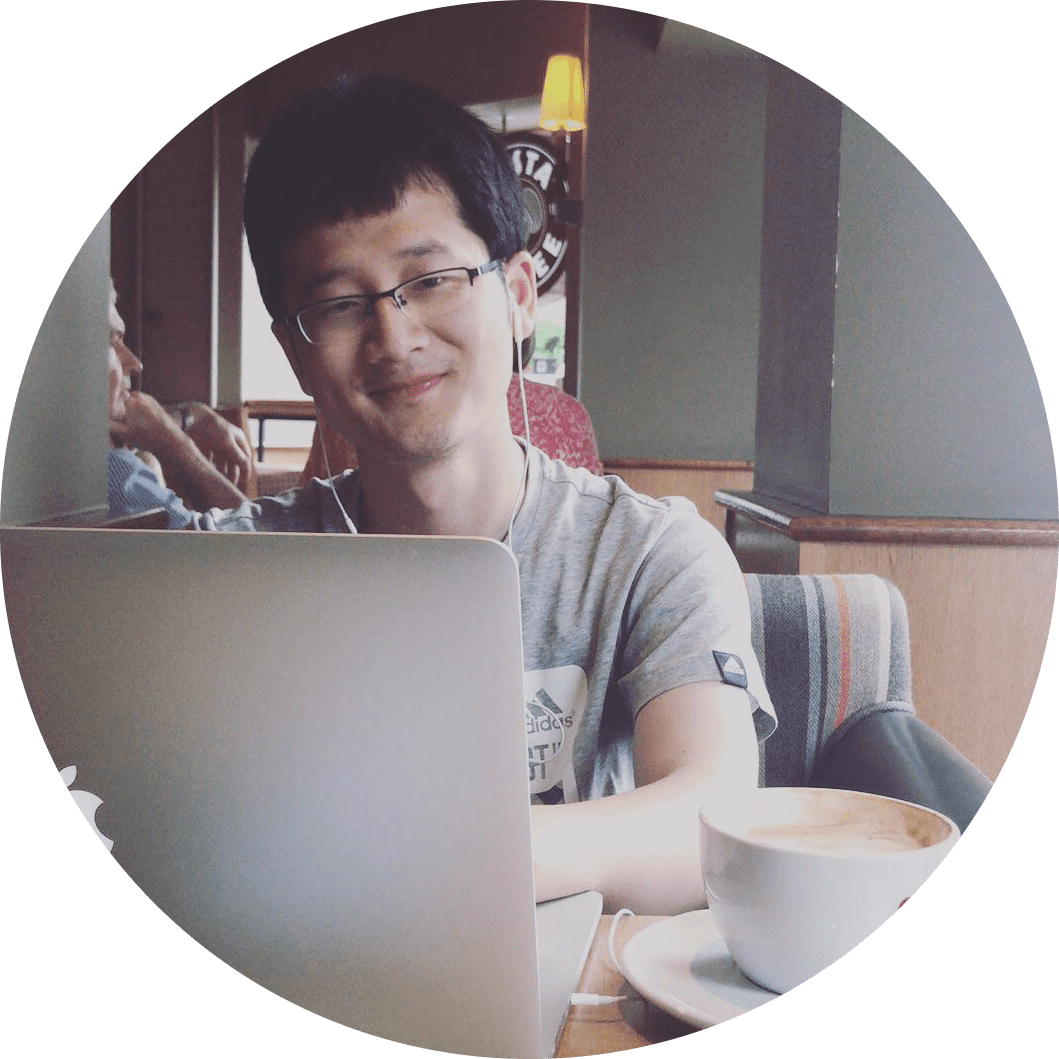}}]{Weidi Xie} is an Associate Professor at Shanghai Jiao Tong University. Weidi completed DPhil at Visual Geometry Group (VGG), University of Oxford, 2018. 
He works on self-supervised representation learning and AI4Medicine, where he has published over 80 papers at top conferences or journals, cited over 17,000 times. 
He regularly serves as reviewer for Nature Medicine, Nature Communications, and area chair for CV conferences, for example, CVPR, NeurIPS, ECCV.

\end{IEEEbiography}

\begin{IEEEbiography}[{\includegraphics[width=1in,height=1.25in,clip,keepaspectratio]{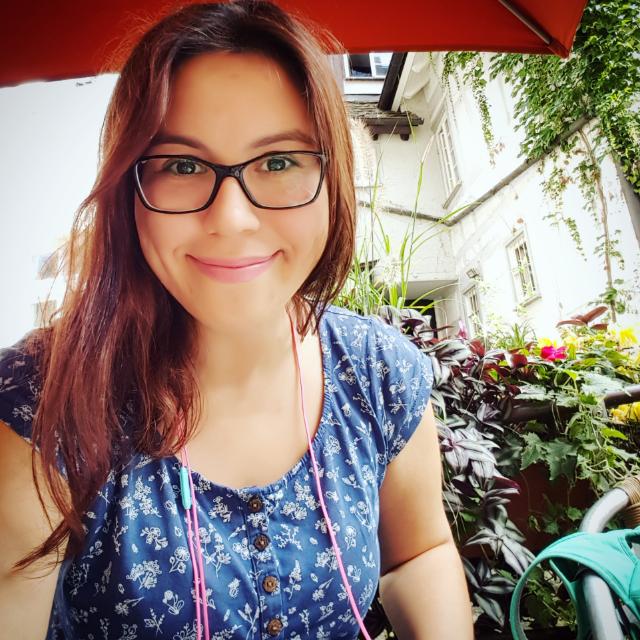}}]{Fatma Güney} is an Assistant Professor at Koç University. She received her Ph.D. in 2017 from MPI and the University of Tübingen. Her research focuses on autonomous driving and 3D vision, with particular interests in motion understanding and uncertainty estimation. Her work has been supported by TÜBİTAK, the European Research Council, and the Royal Society. She regularly serves as a reviewer and area chair at leading CV and ML conferences, including ICCV, ECCV, CVPR, and NeurIPS.

\end{IEEEbiography}